\documentclass[journal]{IEEEtran}

\ifCLASSINFOpdf
   \usepackage[pdftex]{graphicx}
\else
   \usepackage[dvips]{graphicx}
\fi
\usepackage{amssymb, amsmath, amsthm, amsfonts}
\usepackage{breqn}
\newcommand{\pushcode}{\parbox[t]{\dimexpr\linewidth-\algorithmicindent}}

\ifCLASSOPTIONcompsoc
  \usepackage[caption=false,font=normalsize,labelfont=sf,textfont=sf]{subfig}
\else
  \usepackage[caption=false,font=footnotesize]{subfig}
\fi
\usepackage{bm}
\usepackage[noend]{algpseudocode}
\usepackage{tikz}
\usetikzlibrary{patterns}
\usetikzlibrary{chains}

\makeatletter
\DeclareRobustCommand{\rvdots}{%
	\vbox{
		\baselineskip4\p@\lineskiplimit\z@
		\kern-\p@
		\hbox{.}\hbox{.}\hbox{.}
	}}
\makeatother

\usepackage{url}

\usepackage{standalone}

\begin{document}

%
\title{IEEE Copyright Notice}

%
%
%

\author{ 
}

\maketitle



%
\IEEEpeerreviewmaketitle

Copyright \copyright2017 IEEE

Personal use of this material is permitted. Permission from IEEE must be obtained for all other uses, in any current or future media, including reprinting/republishing  this material for advertising or promotional purposes, creating new collective works, for resale or redistribution to servers or  lists, or reuse of any copyrighted component of this work in other works. 
 
Published in: IEEE Transactions on Neural Networks and Learning Systems
\\
URL: \url{http://ieeexplore.ieee.org/document/8004500} \\
DOI: 10.1109/TNNLS.2017.2728818

%
\title{Dimensionality Reduction using Similarity-induced Embeddings}
%
%
%

\author{Nikolaos~Passalis~and~Anastasios~Tefas 
\thanks{Nikolaos Passalis and Anastasios Tefas are with the Department of
		Informatics, Aristotle University of Thessaloniki, Thessaloniki 54124,
		Greece. email: passalis@csd.auth.gr, tefas@aiia.csd.auth.gr}
}

\IEEEoverridecommandlockouts
\IEEEpubid{\makebox[\columnwidth]{DOI 10.1109/TNNLS.2017.2728818 \copyright2017
		IEEE \hfill} \hspace{\columnsep}\makebox[\columnwidth]{ }}

\maketitle

\begin{abstract}
The vast majority of Dimensionality Reduction (DR) techniques rely on second-order statistics to define their optimization objective.  Even though this provides adequate results in most cases, it comes with several shortcomings. The methods require carefully designed  regularizers and they are usually prone to outliers. In this work, a new DR framework, that can directly model the target distribution using the notion of similarity instead of distance, is introduced. The proposed framework, called Similarity Embedding Framework, can overcome the aforementioned limitations and provides a conceptually simpler way to express optimization targets similar to existing DR techniques. Deriving a new  DR technique using the Similarity Embedding Framework becomes simply a matter of choosing an appropriate target similarity matrix. A variety of classical tasks, such as performing supervised dimensionality reduction and providing out-of-of-sample extensions, as well as, new novel techniques, such as providing fast linear embeddings for complex techniques, are demonstrated in this paper using the proposed framework. Six datasets from a diverse range of domains are used to evaluate the proposed method and it is demonstrated that it can outperform many existing DR techniques.
\end{abstract}

\begin{IEEEkeywords}
Dimensionality Reduction, Similarity-based Learning, Kernel Methods
\end{IEEEkeywords}

%
\IEEEpeerreviewmaketitle


\section{Introduction}

Most of the real world data, such as images, videos, audio and text, are inherently high dimensional. To cope with the high dimensionality of the data, many machine learning and data mining techniques employ a preprocessing step that reduces the dimensionality of the data, commonly referred as \textit{Dimensionality Reduction} (DR) \cite{braun2007denoising}. Several DR techniques have been proposed, ranging from classical linear DR methods, such PCA~\cite{pca}, and LDA~\cite{lda}, to non-linear manifold-based techniques, such as Laplacian Eigenmaps~\cite{belkin2001laplacian}, and \mbox{t-SNE}~\cite{van2008visualizing}. Apart from reducing the complexity of the learning problem, DR techniques also dampen the curse of dimensionality \cite{indyk1998approximate}, that affects many machine learning models and allows for better generalization. A DR technique can be formally defined as a mapping $f: \mathbb{R}^d \rightarrow \mathbb{R}^m$ that transforms the input space $\mathbb{R}^d$ into a low dimensional space $\mathbb{R}^m$. The dimensionality of the latter space is less than the dimensionality of the input space, i.e., $m < d$. The low dimensional space is learned by minimizing an appropriately defined objective function.

The type of the learning task that is to be accomplished and the nature of the data determine the DR technique that can be used. For example, when supervised information is available, \textit{supervised} DR techniques, such as LDA \cite{lda}, and MFA \cite{yan2007graph}, are used. On the other hand, when the data lie on a low dimensional manifold embedded in a high dimensional space, \textit{unsupervised} manifold-based techniques, such as Laplacian Eigenmaps~\cite{belkin2001laplacian}, and ISOMAP~\cite{isomap}, can be employed.  \textit{Semi-supervised} methods \cite{belkin2004semi, huang2012semi}, are used when supervised information is available only for some of the data. When linear techniques cannot sufficiently model the data, kernel methods, such as \cite{kpca}, \cite{kda-1}, \cite{kda-2}, are used. Kernel methods, that utilize the so-called \textit{kernel trick} \cite{smola1998learning}, allow simple linear techniques to learn more expressive non-linear mappings.

However, different methods suffer from different shortcomings. For example, LDA cannot learn  a mapping with more dimensions than the number of the training classes (minus one) and tends to overfit the data. Both can negatively affect the quality of the learned mapping. Also, most of the non-linear manifold-based techniques require an extra, non-straightforward learning step to represent data points that were not seen during the training. This process is called out-of-sample extension \cite{gisbrecht2015parametric, xiang2009embedding}. Kernel techniques require the computationally intensive calculation of the kernel matrix to project a point to the learned low-dimensional space and, if the used kernel is not appropriately tuned, they are also prone to overfitting. 

Even though a large number of different DR techniques exist, many of them share a common property: the optimization objective is expressed as a linear combination of the pairwise distances in the low dimensional space. This observation was utilized in the Graph Embedding Framework \cite{yan2007graph}, to unify many existing DR techniques under a common optimization scheme. More specifically, the Graph Embedding Framework defines the following generic objective:

\begin{equation}
\label{eq:graph-embedding}
J_{GE} = \sum_{i=1}^{N}\sum_{j=1, j \neq i}^{N} ||\mathbf{y}_i - \mathbf{y}_j||_2^2 [\mathbf{W}]_{ij} \ \text{s.t.} \ \mathbf{y}^T \mathbf{B} \mathbf{y} = d
\end{equation}
where $d$ is a constant, $\mathbf{y}_i$ ($i=1...N$) is the representation of the points in the low dimensional space, $||\mathbf{y}||_2$ denotes the $l^{2}$ norm of vector $\mathbf{y}$ and the matrices $\mathbf{W}$ and $\mathbf{B}$ allow to express different optimization targets, i.e., different DR methods. The choice of a distance metric, e.g., the euclidean distance, to define the optimization objective is not an arbitrary one: the distance between two points is a natural metric of their relatedness. This fact is also exploited in many machine learning techniques, e.g., k-means, NCC, k-NN and many others \cite{tibshirani2002diagnosis, james2013introduction}. 

However, directly using a distance metric to express the optimization objective, as in (\ref{eq:graph-embedding}), comes with some drawbacks. First, the distance metrics are unbounded, i.e.,  they can range anywhere between 0 and +$\infty$. Therefore, careful regularization is required to ensure that the learned mapping is meaningful. In the Graph Embedding Framework, the regularization is enforced by appropriately setting the matrix $\mathbf{B}$ in  (\ref{eq:graph-embedding}). Also, the optimization is prone to outliers, since disproportionately increasing (or decreasing) the distance of some pairs of points can have significant impact on the objective function. Furthermore, directly manipulating the pairwise distances, might be limiting for some tasks. For example, the Graph Embedding Framework, as well as most of the widely used DR techniques, only utilize second-order statistics to model the target distribution of the data. This limitation is acknowledged by the authors of the Graph Embedding Framework \cite{yan2007graph}. Also, it is not always straightforward to ``clone'' an existing technique (e.g., for providing out-of-sample extensions) using the formulation given in (\ref{eq:graph-embedding}). The importance of defining alternative \textit{similarity metrics}, is best demonstrated in the case of the \mbox{t-SNE}  algorithm~\cite{van2008visualizing}, which is used with great success for visualization tasks. The t-SNE algorithm defines an objective function by first transforming the distances into probabilities using a non-linear mapping. That allows for efficiently tackling the crowding problem that occurs in low-dimensional projections. Therefore, it can be argued that using a similarity metric instead of an unbounded distance metric, can solve some of the aforementioned problems and lead to novel and more accurate DR techniques.

A similarity metric $S$, which is also called \textit{similarity function} through this work, is defined as a function that expresses the affinity between two points  $\mathbf{a}$ and $\mathbf{b}$. The similarity function is bounded to the unit interval, i.e., $0 \leq S(a, b) \leq 1$. The use of a similarity metric, instead of a distance metric, for defining DR techniques can be further justified by the probabilistic interpretation of the notion of similarity. One way to model the probability density function (PDF) $f_{PDF}$ of an unknown distribution is using the Kernel Density Estimation (KDE) method \cite{parzen1962estimation}, also known as Parzen-Rosenblatt method. Given a sample of $N$ points the PDF is estimated as:
\begin{equation}
\label{eq:kde}
\hat{f}_{PDF}(\mathbf{x}) = \frac{1}{N} \sum_{i=1}^N K_{x_i, h}(\mathbf{x})  = \frac{1}{Nh} \sum_{i=1}^N K(\frac{\mathbf{x}-\mathbf{x}_i}{h})
\end{equation}

where $K(\cdot)$ is a kernel, i.e., a function with zero mean that sums to 1, and $h>0$ a scaling parameter that defines the bandwidth of the kernel. Using the KDE method an unknown multi-modal distribution can be modeled without any assumptions regarding the probability distribution. 

A similarity metric acts like a kernel function and allows for modeling the distribution of the data. To understand this note that a similarity function $S_x(\mathbf{y})=S(\mathbf{x}, \mathbf{y})$ provides the PDF induced by the point $\mathbf{x}$. Therefore, summing over the similarities induced by the sampled points (similarly to (\ref{eq:kde})) provides a similarity-based way to directly define the PDF of the data, without relying on second-order statistics, like in (\ref{eq:graph-embedding}). This fact is further explained in Section \ref{section:proposed-method}, where the proposed method is presented.

In this work, a novel DR framework that defines a generic optimization objective using the notion of {similarity}, instead of distance, is proposed.  Using the proposed framework, which is called Similarity Embedding Framework (SEF), different DR techniques can be derived by simply setting the appropriate target similarity matrix. In other words, the proposed framework allows for embedding the data in a space that follows a predefined target PDF. By choosing different target PDFs, different techniques can be derived. Apart from deriving methods inspired from existing techniques, such as the PCA, LDA, Laplacian Eigenmaps, etc., the proposed framework can serve as a basis for creating new DR techniques. Several novel ways to use the proposed framework are provided in this paper. Note that any type of DR methods, i.e., unsupervised, supervised or semi-supervised, can be derived within the SEF.  The SEF can be combined with any (differentiable) model, such as a linear projection function, a kernel-based projection function, or even a deep neural network, to perform the mapping between the input space and the low dimensional space. Also, different options exist to define the pairwise similarities. In this work, a non-linear transformation of a pairwise distance metric into a bounded similarity measure using a Gaussian Kernel is utilized. Even though the similarity between two points is a function of their distance, when a non-linear transformation is used different solutions to the optimization problem can be obtained (the distances are not evenly transformed, the effect of the outliers is lessen, etc.).

The contributions of this work are summarized as follows: 
\begin{itemize}
	\item A generic framework for similarity-based DR is proposed. Two different mapping functions are considered for projecting the initial space into a low dimensional space. To the best of our knowledge, this is the most general framework for DR, since it can be used in order to derive completely new methods, it can repurpose classification methods (e.g., SVM) in order to provide novel DR techniques and it can clone almost any existing DR method, while overcoming some of their limitations
	\item It is demonstrated how to derive optimization targets inspired from a wide range of existing techniques. This allows for overcoming the original limitations of some methods, such as learning LDA-like embeddings with an arbitrary number of dimensions. A method for LDA-like projections is derived and evaluated.
	\item A method for ``cloning'' the PDF of existing techniques is proposed and evaluated. That allows for providing out-of-sample extensions, as well as, fast linear approximations of existing techniques, such as kernel-based techniques or methods that produce high-dimensional spaces.
	\item A novel technique, that is based on the proposed framework, for learning SVM-based projections, that can overcome the need for learning separate binary SVM classifiers, is proposed. The learned projection can be combined with a simple and fast classifier, such as the Nearest Class Centroid (NCC) \cite{tibshirani2002diagnosis}, and greatly reduces the classification time with little effect on the classification accuracy.
\end{itemize}

The rest of the paper is structured as follows. The related work is discussed in Section \ref{section:related-work}. Then, the proposed framework is introduced in Section \ref{section:proposed-method}, while different ways to set the target similarity matrix (supplemented with a set of toy examples) are presented in Subsection \ref{section:optimization-targets}. 
 The experimental evaluation of the proposed methods using six different datasets from a diverse range of domains is provided in Section \ref{section:experiments}. Finally, conclusions are drawn and future research directions are discussed in Section \ref{section:conclusions}. A reference open-source implementation of the proposed framework is available at \url{https://github.com/passalis/sef}.

\section{Related Work}
\label{section:related-work}

The DR techniques can be roughly categorized into two groups. The first category is composed of those methods that use a (linear or non-linear) projection function to map the input space into the low dimensional space. Typical examples of linear projection techniques are the PCA technique \cite{pca}, which learns a projection that maximizes the variance of the data in the low dimensions space, and the LDA technique \cite{lda}, which learns a projection that minimizes the within-class scatter and maximizes the between-class scatter. Several extensions have been proposed, including kernelized versions \cite{kpca, kda-1, kda-2}, and more regularized  versions \cite{yan2007graph, laplacian-pca}. Another popular technique, that learns generative models of the data, is based on autoencoders \cite{baldi2012autoencoders}. Again, several extensions have been proposed, such as denoising autoencoders \cite{vincent2010stacked}, that learn models more robust to noise, and sparse autoencoders \cite{deng2013sparse}, that include sparsity constraints to learn sparse representation of the data. 
The proposed framework provides a simplified and general way to express optimization targets inspired by many of these techniques by simply setting an appropriate target similarity matrix, while overcoming some of their original limitations. This is further discussed in Subsection \ref{section:optimization-targets}, where different ways to choose the optimization target are considered.

The other category of techniques only learn the low dimensional representation of the training data, instead of learning an explicit mapping between the input and the output space.  This greatly restricts the applicability of these techniques, when there is the need of representing points that were not included in the original training set. Among the most widely used such techniques are the ISOMAP \cite{isomap}, which seeks an embedding that maintains the geodesic distances of the points, and the Laplacian Eigenmaps \cite{belkin2001laplacian}, that seeks an embedding where the data that are close in the original space are also close in the low-dimensional space. The optimization targets of most embedding based techniques can be also expressed under the proposed framework, leading to their similarity-based counterparts.

Several methods for providing out-of-sample extensions were developed to overcome the aforementioned limitation. The first category of the out-of-sample extensions includes methods that train a projection using the same objective function that is utilized in the corresponding method \cite{maaten2009learning, law2006incremental, bunte2012general}. However, these methods must be appropriately modified to provide out-of-sample extensions for different techniques. This problem is addressed in the interpolation/regression-based techniques that are able to provide general out-of-sample extensions for any method \cite{gisbrecht2015parametric, xiang2009embedding}. The framework proposed in this work provides a novel way to produce out-of-sample extensions by performing regression of the pairwise similarity of the training samples, instead of the actual representation of the data. It was experimentally established that this leads to better generalization ability.

To the best of our knowledge, the SEF is the only framework that expresses the process of dimensionality reduction using the notion of similarity and it is able to derive several well-known DR objectives by simply setting an appropriately chosen target similarity matrix. In contrast to the existing frameworks, such as \cite{yan2007graph} and \cite{kokiopoulou2011trace}, that define the optimization objective by directly using the pairwise distances, the SEF uses a non-linear transformation of the distances to define the optimization objective. This allows for overcoming several shortcomings of existing techniques and provides a conceptually easier way to derive new DR techniques. Furthermore, the proposed approach is also different from the Spectral Regression Framework \cite{cai2007spectral}, that reduces the problem of DR into a regression problem. The SEF moves beyond this approach, since it does not directly perform regression on the actual representation of the data, which constraints the method into the given representation targets. Instead, it uses similarity targets and allows for learning any representation that satisfy the given similarity constraints.

It should be noted that some methods, such as the t-SNE \cite{van2008visualizing}, and the Simbed \cite{lee2009simbed}, also utilize a similarity-like metric. However, these methods solve a specific problem and do not provide a general DR framework. Actually, as shown in Subsection~\ref{section:optimization-targets}, the t-SNE algorithm can be derived as a special case of the proposed framework. Similar results can be also derived for the other similarity-based techniques, e.g.,  the Simbed \cite{lee2009simbed}.

\section{Similarity Embedding Framework}
\label{section:proposed-method}

In this Section the Similarity Embedding Framework is presented. First, the used notation and the general formulation of the framework are introduced. Then, two projection functions, a linear function and a kernel-based function, are considered and a learning algorithm for the proposed framework is described. Finally, it is demonstrated how to derive methods inspired from existing techniques, as well as, new novel methods by appropriately setting the target  similarity matrix.

\subsection{Similarity Embedding Framework}

Let $\mathcal{X}_{train}=\{\mathbf{x}_1, \mathbf{x}_2, ..., \mathbf{x}_N\}$ be a collection of data points, where  $\mathbf{x}_i \in \mathbb{R}^d$ and $d$ is the original dimensionality of the data. The proposed method aims to learn an embedding function $f_W: \mathbb{R}^d \rightarrow \mathbb{R}^m$ that projects the data of  $\mathcal{X}_{train}$ to a lower dimensional space ($m < d$), where the similarity between each pair of data points is transformed according to a predefined target. The domain of the mapping function $f_W$ can be either restricted to the set $\mathcal{X}_{train}$, i.e., provide a direct embedding, or extend to the whole feature space, i.e. provide a projection. It is reminded that  $S(\mathbf{x}_i, \mathbf{x}_j)$ is a function that measures the similarity between two data points $\mathbf{x}_i$ and $\mathbf{x}_j$. The similarity matrix of the embedded data is defined as: $[\mathbf{P}]_{ij} = S(f_W(\mathbf{x}_i), f_W(\mathbf{x}_j))$, where the notation $[\mathbf{P}]_{ij}$ is used to denote the element in the $i$-th row and the $j$-th column of the matrix $\mathbf{P}$. The SEF aims to learn an embedding that makes the similarities in the projected space as similar as possible to a predefined target.
Any source of information, e.g., supervised information (labels), manipulations of the original similarity of the data, non-parametric dimensionality reduction techniques, etc., can be used to define the target similarity matrix $\mathbf{T} \in \mathbb{R}^{n\times n}$. Each matrix element $[\mathbf{T}]_{ij}$ contains the desired similarity between the $i$-th and the $j$-th training samples in the lower dimensional space. For example, to perform supervised DR one can set the target similarity between samples of the same class to 1, while zeroing the target similarity between samples that belong to different classes. By adopting the probabilistic interpretation of the proposed framework, the target similarity matrix $\mathbf{T}$ defines the target PDF, while matrix $\mathbf{P}$ is used to model the PDF of the learned embedding that must closely resemble the target PDF.

In order to enforce the target similarity in the lower-dimensional space, a loss function must be defined. In this work, the following objective function is used for the optimization of $f_W$:
\begin{equation}
\label{eq:objective}
J_s = \frac{1}{2 ||\mathbf{M}||_1}\sum_{i=1}^{N} \sum_{j=1}^{N} [\mathbf{M}]_{ij}([\mathbf{P}]_{ij}-[\mathbf{T}]_{ij})^2
\end{equation}
where $\mathbf{M} \in \mathbb{R}^{n \times n}$ is a weighting mask that defines the  importance of achieving the target similarity between two points in the projected space. The values of the weighting masks are restricted to the unit interval, i.e., $0 \leq [\mathbf{M}]_{ij} \leq 1$. The notation $||\mathbf{M}||_1$ is used to refer to the element-wise 1-norm of the matrix $\mathbf{M}$, i.e., $||\mathbf{M}||_1 = \sum_{i=0}^N \sum_{j=0}^N |[\mathbf{M}]_{ij}|$.  The objective function (\ref{eq:objective}) is minimized when each pair of the projected points achieves its target similarity. When $ \forall i,j: [\mathbf{M}]_{ij}=1$, then minimizing the objective function is equivalent to minimizing the Frobenius norm of the difference between the projection similarity matrix $\mathbf{P}$ and the target similarity matrix $\mathbf{T}$. Note that the specified objective function penalizes each pair that has different (either higher or lower) similarity from its target. This is in contrast to the Graph Embedding Framework, where complex combinations between the optimization objective and the regularizer must be used to penalize both cases. More sophisticated objective functions can be also defined, e.g., based on the divergence between the similarities of the embedding and the target. However, experimentally it was established that the proposed objective function works quite well, while allowing for easier and faster implementation of the proposed framework.

Any differentiable function $S(\mathbf{x}_i, \mathbf{x}_j)$ can be used to measure the similarity of the projected data. In this work, the Gaussian kernel, also known as Heat kernel, combined with the euclidean distance is used as similarity measure: $ S(\mathbf{x}_i, \mathbf{x}_j)) = exp (-||\mathbf{x}_i -  \mathbf{x}_j||^2_2 / \sigma_P)$, where $\sigma_P$ is the scaling factor of the similarity function. The scaling factor $\sigma_P$  acts similarly to the bandwidth of the KDE method.  Note that for very large distances, there is little change in the obtained similarity. This non-linear behavior also acts as an intrinsic regularizer that prevents the existence of outliers in the projected space. Using the chosen similarity function the similarity matrix $\mathbf{P}$ is redefined as:
\begin{equation}
\label{eq:similarity-projected}
[\mathbf{P}]_{ij} = exp (-||f_W(\mathbf{x}_i) - f_W(\mathbf{x}_j)||^2_2 / \sigma_P)
\end{equation}

Also, several ways exist to define the projection function $f_W$. The choices range from a simple linear projection, i.e., a linear transformation of the data, to more advanced techniques, such as, kernel projections or deep neural networks. In the next two subsections two different projection functions, a linear and a non-linear, are considered. Since, the gradient descent technique is used to optimize the projection function, the derivative of the objective function with respect to the parameters of each projection is also derived.

Note that when the data are projected to several dimensions it is possible to learn degenerate solutions, such as, learning the same projection for all the learned dimensions. To avoid this behavior, a regularization term $J_{p}$ is added to the projection function that enforces the orthonormality of the projection directions. Therefore, the final objective function is defined as:

\begin{equation}
\label{eq:final_objective}
	J = (2-\alpha_p) J_s + \alpha_p J_p, \ \ 0 \leq \alpha_p \leq 1
\end{equation}

The parameter $\alpha_p$ alters the importance of the orthonormality regularizer (for $\alpha_P=0$ no regularization is used, while for $\alpha_p=1$ the orthonormality is equally important to the objective $J_s$). Note that other regularization methods, such as sparsity regularization or non-negative regularization techniques, can be also used.

\subsection{Linear Similarity Embedding}
The first candidate projection function $f_W$ that is considered is a simple linear transformation of the input space, i.e., $f_W(\mathbf{x}) = \mathbf{W}^T \mathbf{x}$, where $\mathbf{W} \in \mathbb{R}^{d \times m}$ is the projection matrix. To simplify the presentation of the proposed approach, the matrix of the original data $\mathbf{X} = [\mathbf{x}_1, \mathbf{x}_2, ..., \mathbf{x}_N]^T \in \mathbb{R}^{n \times d}$ is defined, as well as, the matrix of the projected data: $\mathbf{Y} = [\mathbf{y}_1, \mathbf{y}_2, ..., \mathbf{y}_N]^T \in \mathbb{R}^{n \times m}$, where $\mathbf{y}_i = f_W(\mathbf{x}_i)$.

The derivative of the objective function with respect to the weights of the linear projection is derived as:
\begin{equation}
\label{eq:lsp-gradient}
\frac{\partial J_s}{\partial [\mathbf{W}]_{kt}} = \frac{1}{||\mathbf{M}||_1} \sum_{i=1}^N \sum_{j=1}^N [\mathbf{M}]_{ij}([\mathbf{P}]_{ij}-[\mathbf{T}]_{ij}) \frac{\partial [\mathbf{P}]_{ij}}{\partial [\mathbf{W}]_{kt}}  
\end{equation}
where $\frac{\partial [\mathbf{P}]_{ij}}{\partial [\mathbf{W}]_{kt}}   = -\frac{2}{\sigma_P}[\mathbf{P}]_{ij} ([\mathbf{Y}]_{it} - [\mathbf{Y}]_{jt} ) ([\mathbf{X}]_{ik} - [\mathbf{X}]_{jk} )$.

Also, the orthonormality regularizer must be specified. For the linear projection it is defined as:
\begin{equation*}
	J_p = \frac{1}{2m^2} ||\mathbf{W}^T \mathbf{W} - \mathbf{I}_{m\times m}||^2_F
\end{equation*} 
where $\mathbf{I}_{m\times m}$ is the $m \times m$ identity matrix. 

The derivative of the orthogonality regularizer is calculated as: 
\begin{equation}
\label{eq:ortho-deriative}
	\frac{\partial J_p} {\partial [\mathbf{W}]_i} = \frac{2}{m^2} \sum_{j=1}^{m} ([\mathbf{W}]_i^T [\mathbf{W}]_j - \delta_{ij}) [\mathbf{W}]_j
\end{equation} 
where the notation $[\mathbf{W}]_i$ is used to refer to the $i$-th column vector of the matrix $\mathbf{W}$ and $\delta_{ij}$ is the Kronecker delta function.

The gradient descent algorithm can be used for optimizing the objective (\ref{eq:objective}) using the equations (\ref{eq:lsp-gradient}) and (\ref{eq:ortho-deriative}):
\begin{equation}
\Delta \mathbf{W} = - \eta \frac{\partial J}{\partial \mathbf{W}}
\end{equation}
where $\eta$ is the learning rate. In this work, the Adam algorithm \cite{kingma2014adam}, is used instead of the simple gradient descent, since it provides faster and more stable convergence.

\subsection{Kernel Similarity Embedding}
To define the kernel similarity embedding, the input space is first (non-linearly) transformed into a higher dimensional space and then a linear projection is used to acquire the final low-dimensional space. Let $\phi:\mathcal{X} \rightarrow \mathcal{H}$ be a mapping of the original space $\mathcal{X}$ into a Hilbert space $\mathcal{H}$ of arbitrary (and possible infinite) dimensionality $h$. The matrix of the data in the Hilbert space is defined as $\mathbf{\Phi} = \phi(\mathbf{X}) \in \mathbb{R}^{n\times h}$, where the mapping $\phi(\mathbf{x})$ is applied for every (row) vector $\mathbf{x}_i$ of $\mathbf{X}$. In the kernelized version of the proposed method, a linear transformation of the data in the Hilbert space, instead of the original space, is learned. Note that the weight matrix $\mathbf{W}$ might have infinite dimensions as well, since it is used to project the data from Hilbert space $\mathcal{H}$ into to final space $\mathcal{T}$. 

Directly optimizing the weight matrix in this case is usually impractical (or even not possible). To this end, the Representer theorem \cite{smola1998learning}, is employed to express the weight matrix as a linear combination of the data points $\mathcal{X}_{train}$. Therefore, the matrix $\mathbf{W}$ is redefined as:
\begin{equation}
\label{eq:weights-kernel}
\mathbf{W} = \phi(\mathbf{X})^T \mathbf{A} = \mathbf{\Phi}^T \mathbf{A}
\end{equation}
where $\mathbf{A} \in \mathbb{R}^{n \times m}$ is a coefficient matrix that defines each projection as a linear combination of the data points. The projection can be now calculated, using the Eq. (\ref{eq:weights-kernel}) as: 

\begin{equation*}
\mathbf{Y}^T = \mathbf{W}^T \mathbf{\Phi}^T = \mathbf{A}^T \mathbf{\Phi} \mathbf{\Phi}^T = \mathbf{A}^T\mathbf{K}
\end{equation*}
where $\mathbf{K} = \mathbf{\Phi} \mathbf{\Phi}^T  \in \mathbb{R}^{n \times n}$ is the kernel matrix of the data that contains the inner products between the data points in the Hilbert space, i.e., $[\mathbf{K}]_{ij} = \phi(\mathbf{x}_i)^T\phi(\mathbf{x}_j)$.  When a Reproducing Kernel Hilbert Space (RKHS) is used the kernel matrix can be calculated without explicitly calculating the inner products in the Hilbert space. Therefore, different choices for the kernel matrix $\mathbf{K}$ lead to different Hilbert spaces. Among the most used kernels is the Gaussian/RBF kernel, i.e., $[\mathbf{K}]_{ij} = exp(-||\mathbf{x}_i-\mathbf{x}_j||_2^2 / \gamma^2)$, which maps the input points to an infinite dimensional space. Several other kernels functions have been proposed in the literature, e.g., polynomial kernels \cite{smola1998learning}, and they can be also used with the proposed method.

Therefore, in the kernelized method the coefficient matrix $\mathbf{A}$ is to be learned, instead of the weight matrix $\mathbf{W}$. Again, the gradient descent algorithm is used and the corresponding gradient is derived as:

\begin{equation}
\label{eq:ksp-gradient}
\frac{\partial J}{\partial [\mathbf{A}]_{kt}} = \frac{1}{||\mathbf{M}||_1}  \sum_{i=1}^N \sum_{j=1}^N [\mathbf{M}]_{ij}([\mathbf{P}]_{ij}-[\mathbf{T}]_{ij}) \frac{\partial [\mathbf{P}]_{ij}}{\partial [\mathbf{A}]_{kt}}  
\end{equation}
where $
\frac{\partial [\mathbf{P}]_{ij}}{\partial [\mathbf{A}]_{kt}}   = -\frac{2}{\sigma_P}[\mathbf{P}]_{ij} ([\mathbf{Y}]_{it} - [\mathbf{Y}]_{jt} ) ([\mathbf{K}]_{ik} - [\mathbf{K}]_{jk})
$.

For the kernel projection the orthonormality regularizer is defined as: $
	J_p = \frac{1}{2m^2}||\mathbf{W}^T \mathbf{W} - \mathbf{I}_{m\times m}||^2_F = \frac{1}{2m^2} ||\mathbf{A}^T \mathbf{K} \mathbf{A} - \mathbf{I}_{m\times m}||^2_F$  and the corresponding orthonormality derivative is derived as: $
	\frac{\partial J_p} {\partial [\mathbf{A}]_i} = \frac{2}{m^2} \sum_{j=1}^{m} ([\mathbf{A}]_i^T \mathbf{K} [\mathbf{A}]_j - \delta_{ij}) \mathbf{K} [\mathbf{A}]_j$.

\subsection{Learning Algorithm}

The complete learning algorithm for the liner/kernel similarity embeddings is shown in Figure \ref{algo:learning-algo}. First, the data points are z-normalized, i.e., transformed to have zero mean and unit variance (line~2). Using z-normalization leads to smoother and faster convergence, as it was experimentally established. For the kernel method, the data are only transformed to have zero mean (to avoid instabilities due to the larger values during the kernel matrix calculation). Then, the parameters of the projection are initialized using the projection weights of the PCA method (or the Kernel PCA method, if a kernel projection is used) (line~3). The parameters can be also randomly initialized. However, this can lead to instabilities during the optimization process if the scale of the initial weights is not carefully selected. Also, using PCA-based initialization effectively turns the method into a deterministic one, since the learned embedding is dependent to the used initialization (the optimization problem is not convex). Note that the orthonormality constraint is already satisfied by the initial projection provided by the (linear) PCA method. For the Kernel PCA method, the projection is also re-normalized to ensure that each projection vector has unit norm. This is necessary since the used kernel matrix is not centered, as in the Kernel PCA method.  

\begin{figure}
\includegraphics[width=0.32\linewidth]{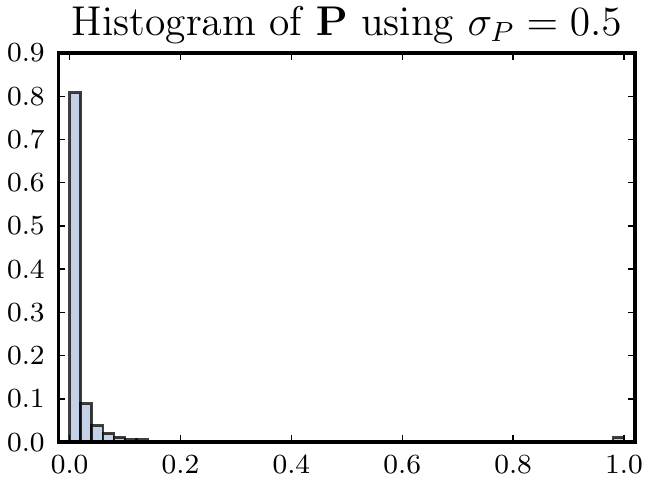}
\includegraphics[width=0.32\linewidth]{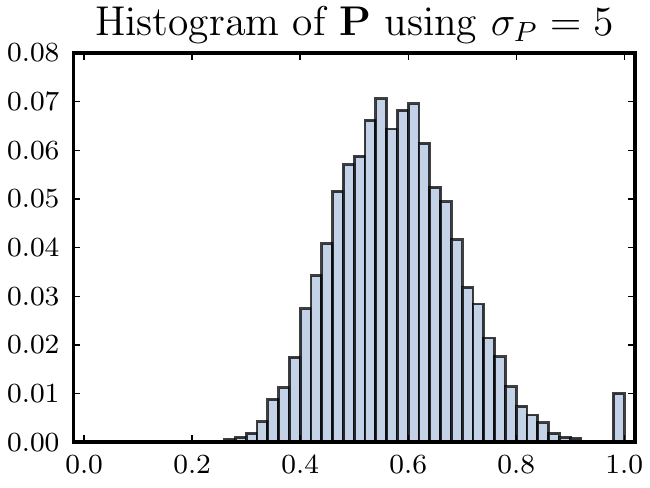}
\includegraphics[width=0.32\linewidth]{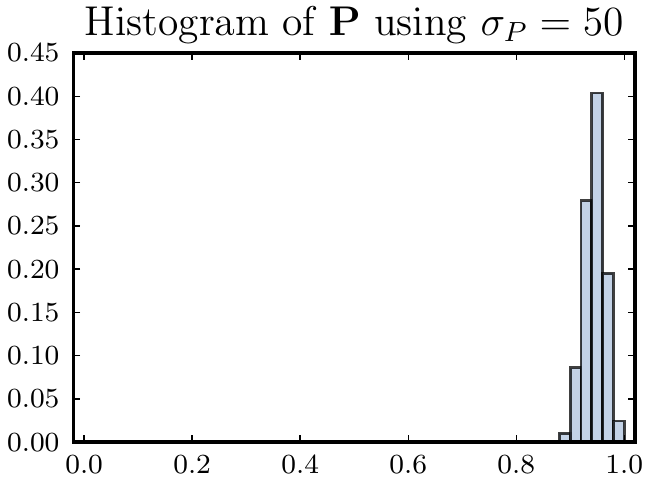}
\caption{Effect of the scaling parameter $\sigma_P$ on the distribution of the similarity values}
\label{fig:similarity-distribution-projection}
\end{figure}

Then, the scaling factor $\sigma_P$ is selected (lines 9-16). The optimal value for the scaling factor $\sigma_P$ depends on the initialization of the parameters of the projection. If the value of $\sigma_P$ is too small, then all the similarity values of the projection tend to zero. On the other hand, if the value of $\sigma_P$ is very large, all the projected data points become similar to each other and the similarity values tend to 1. Either cases must be avoided. The effect of different values of $\sigma_P$ on the distribution of the similarities is illustrated in Figure \ref{fig:similarity-distribution-projection}. Experimentally, it was established that values of $\sigma_P$ that ``spread'' the similarity values as much as possible allows for faster convergence. In this work, a simple heuristic criterion combined with line search is used to select the value of $\sigma_P$ (lines 12-15). First, the histogram of the values of $P$ is computed using 100 bins. Then, the value of $\sigma_P$ that minimizes the maximum value of the histogram of $\mathbf{P}$ is selected. The candidate values for the $\sigma_P$ range from $10^{-5}$ to $10^{5}$.  

Finally, the parameters of the projection are optimized using the Adam algorithm (lines 5-7). The default hyper-parameters are used for the Adam algorithm. The learning rate is fixed to $\eta = 10^{-3}$ for the linear method and to $\eta = 10^{-5}$ for the kernel method.

\begin{figure}

	\hrule
	\textbf{Input:} A set $\mathcal{X}_{train}=\{\mathit{x}_1, ..., \mathit{x}_N\}$ of $N$ training points and a target similarity matrix $\mathbf{T}$.\\
	\textbf{Hyper-parameters}:  $N_{iters}$, $\eta$,  $\gamma$ (for kernel method only) \\
	\textbf{Output}: The parameters of the projection
	\hrule
	\begin{algorithmic}[1]
		
		\Procedure{SimilarityEmbeddingLearning}{}
		\State Scale the data using z-normalization
		\State \pushcode{Use the PCA/KPCA technique to initialize the parameters ($\mathbf{W}$ for the linear method, $\mathbf{A}$ for the kernel method) of the projection.}
		\vspace*{-2px}
		\State $\sigma_P=\textit{ScalingFactorEstimation()}$		
		\For {$i\gets 1; i\leq N_{iters}; i++$}
		\State \pushcode {Calculate the parameter's derivative using Eq. (\ref{eq:lsp-gradient}) \\ for the linear method or Eq. (\ref{eq:ksp-gradient}) for the kernel \\method}
		\vspace{0.1px}
		\State \pushcode{Apply the Adam algorithm using learning rate $\eta$ \\ to update the projection's parameters} 
		\EndFor \\
		\Return $\mathbf{W}$ for the linear method, $\mathbf{A}$ for the kernel method

		\EndProcedure
		\Procedure{ScalingFactorEstimation}{}
		\State $best_{\sigma_P} \leftarrow 0$, $minmax \leftarrow \infty$
		\For {each candidate $\sigma_P$}
		\State Calculate the similarity matrix $\mathbf{P}$ using Eq.
		 (\ref{eq:similarity-projected})
		\If {$\max(histogram(P, 100)) < minmax$}
		\State $minmax \leftarrow \max(histogram(P, 100))$
		\State $best_{\sigma_P} \leftarrow \sigma$
		\EndIf
		\EndFor		
		\Return  $best_{\sigma_P}$
		\EndProcedure
\end{algorithmic}
	\hrule
	\caption{Similarity Embedding Learning Algorithm}
	\label{algo:learning-algo}
\end{figure}

The iterative nature of the SEF learning algorithm is expected to allow it scale well with large datasets. However, calculating large similarity matrices is limiting for large-scale datasets. To address this problem, the proposed learning technique can be extended to allow incremental learning by using only a subset of the similarity matrix for each iteration. During the testing, the computational complexity depends only on the used projection function. The linear projection function is very fast, requiring only the calculation of one product between the projection matrix and an  input sample. This allows for providing fast linear embeddings for a wide range of techniques (Sections \ref{section:out-of-sample} and \ref{section:experiments-fast}). The kernel projection function is more computationally intensive requiring the calculation of the kernel function between an input sample and all the training samples. However, the testing time for the kernel projection can be reduced by using sparsity constraints and, as a result, using only a subset of the training samples to project a new sample.

The proposed methods were implemented using the Theano library \cite{2016arXiv160502688short}. Theano also allows for easily optimizing the model using the GPU, instead of the CPU, significantly accelerating the learning process. Learning a 50-dimensional projection using 1000 training samples and 500 training iterations requires less than 5 seconds on an entry level GPU, compared to 30 seconds for a 4-core CPU.

\subsection{SEF Target Similarity Matrices}
\label{section:optimization-targets}

There are several ways to select the target similarity matrix $\mathbf{T}$, with each one leading to a different DR technique. First, it is demonstrated how to acquire optimization targets similar to existing DR techniques. Next, a method for ``cloning'' any existing technique, without any prior knowledge, is provided. This can be used to either provide out-of-sample extensions or fast approximation for computationally intensive techniques. Finally, a novel way to perform SVM-based supervised dimensionality reduction is proposed.

\subsubsection{Existing Techniques Derivation}
In this subsection it is demonstrated how to set optimization targets inspired by existing techniques. It should be noted that the derived techniques are not strictly equivalent to the existing approaches, mainly due to the non-linear scaling of the distances using the Heat kernel~(\ref{eq:similarity-projected}). This allows for overcoming some of the original limitations of the techniques and to achieve better classification results using less dimensions. Note that the scope of this section is to provide the intuition behind using the SEF to derive existing techniques and not to provide an exhausting list of techniques that can be expressed using the SEF. 

The following notation is used in the rest of the paper: Let \textit{XYZ} be a DR technique. The linear similarity embedding using a target similarity matrix inspired by \textit{XYZ} is denoted by \mbox{\textit{S-XYZ}}, while the kernel similarity embedding by \textit{\mbox{KS-XYZ}}. The kernel technique is combined with an RBF kernel and, unless otherwise stated, the optimization mask is set to 1 ($[\mathbf{M}]_{ij}=1$).

To assist the presentation of the techniques several toy examples are provided. To this end, 500 images from 3 classes (digits 0, 1 and 2) of the MNIST dataset are used \cite{lecun1998mnist}. The optimization runs for 1000 iterations for all the experiments, unless otherwise stated. More figures, along with a comparison with other techniques, are provided in the supplementary material.

\begin{figure*}
\begin{center}
\subfloat[]{\includegraphics[width=0.19\linewidth]{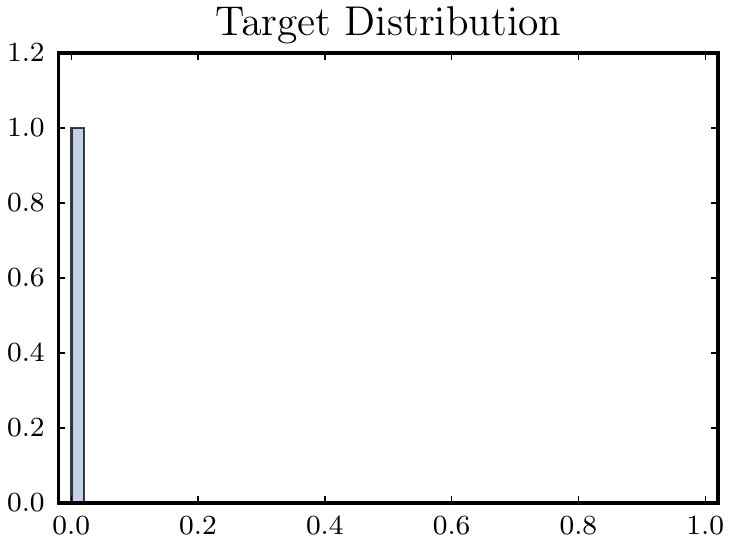}
}
\subfloat[]{\includegraphics[width=0.19\linewidth]{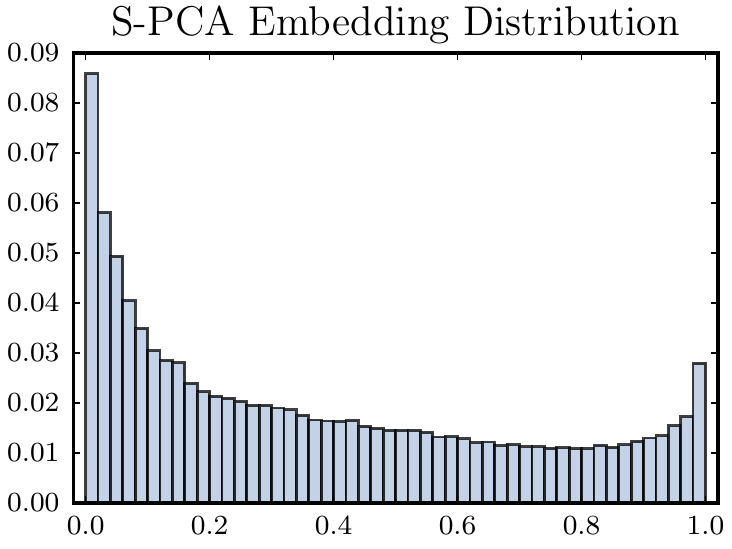}}
\subfloat[]{\includegraphics[width=0.19\linewidth]{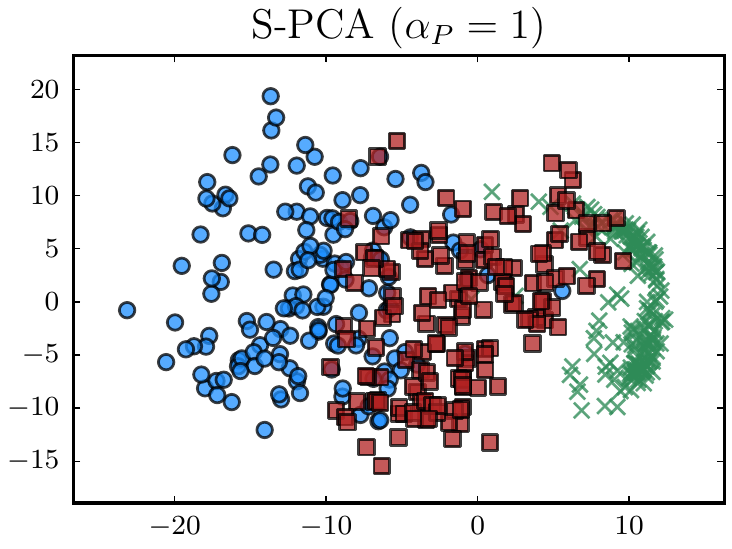}}
\subfloat[]{\includegraphics[width=0.19\linewidth]{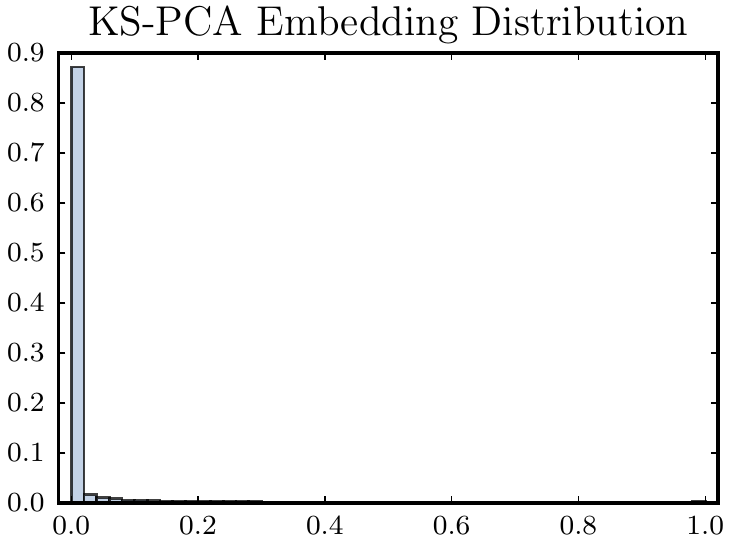}}
\subfloat[]{\includegraphics[width=0.19\linewidth]{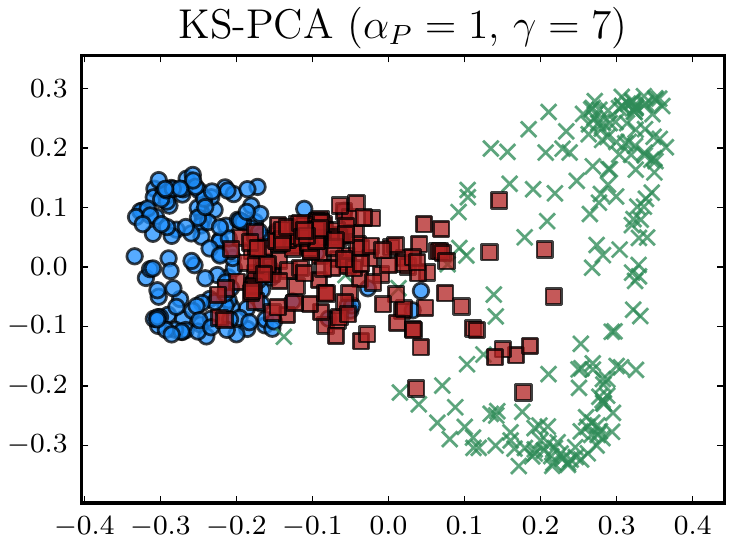}}

\subfloat[]{\includegraphics[width=0.19\linewidth]{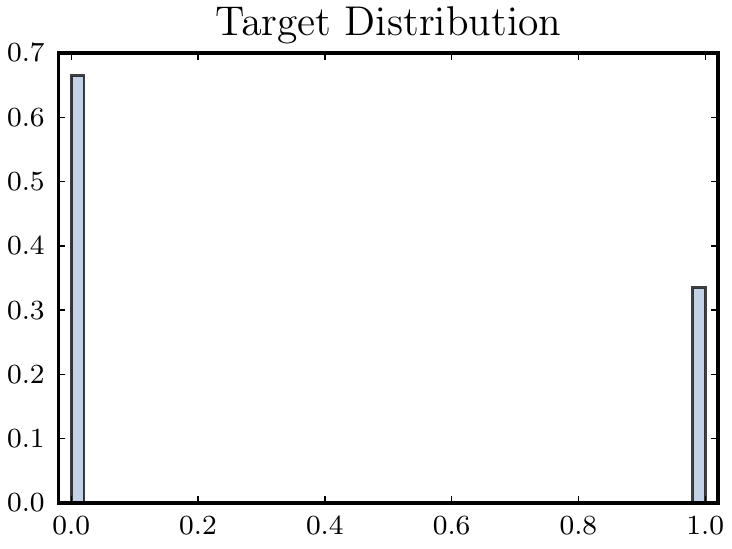}}
\subfloat[]{\includegraphics[width=0.19\linewidth]{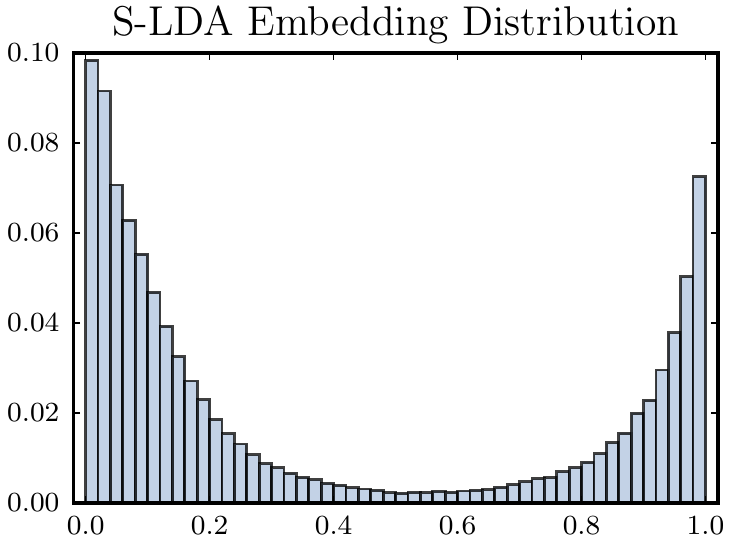}}
\subfloat[]{\includegraphics[width=0.19\linewidth]{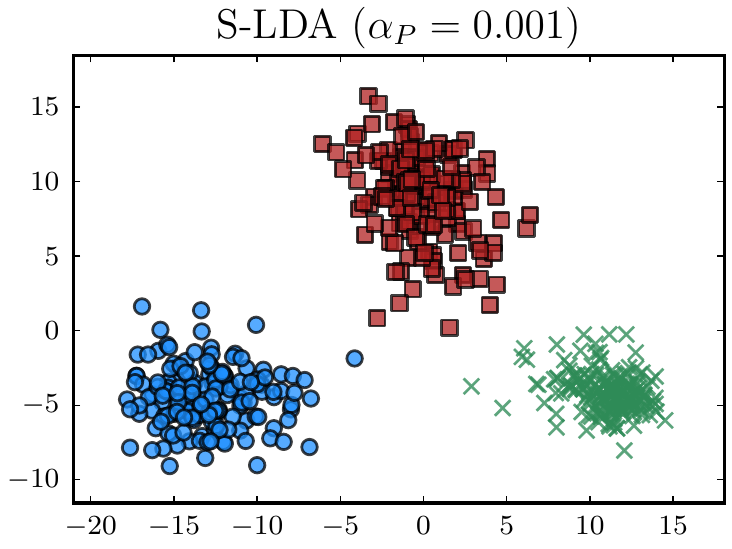}}
\subfloat[]{\includegraphics[width=0.19\linewidth]{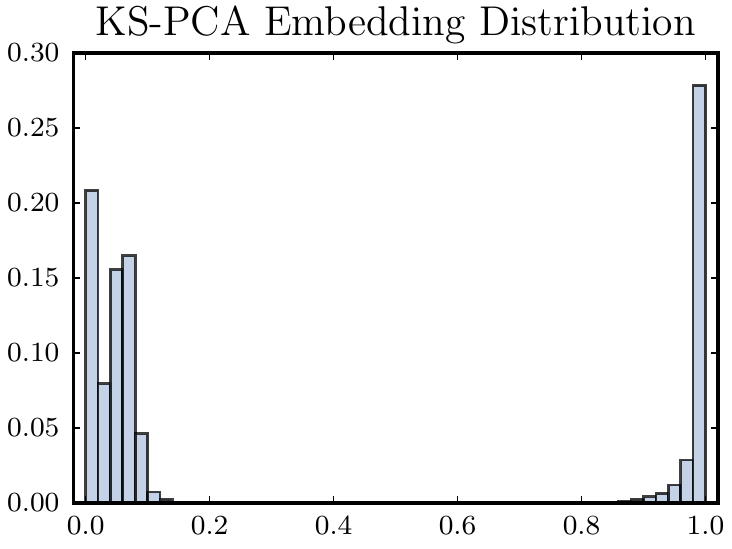}}
\subfloat[]{\includegraphics[width=0.19\linewidth]{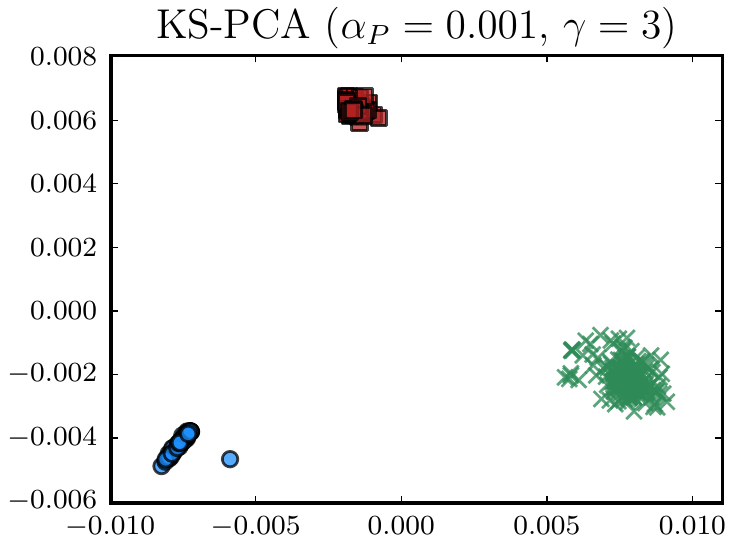}}

\subfloat[]{\includegraphics[width=0.19\linewidth]{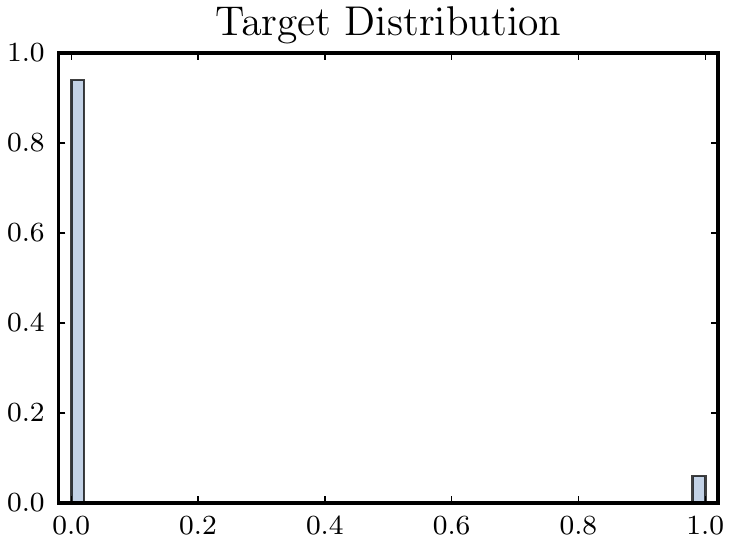}}
\subfloat[]{\includegraphics[width=0.19\linewidth]{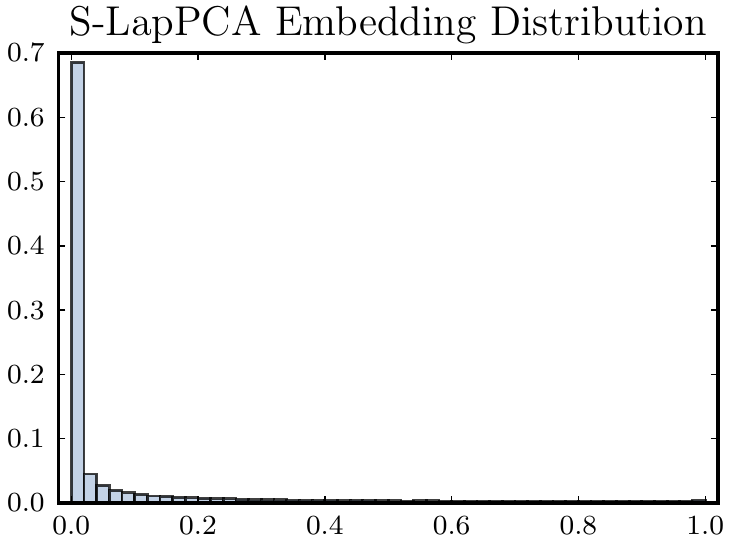}}
\subfloat[]{\includegraphics[width=0.19\linewidth]{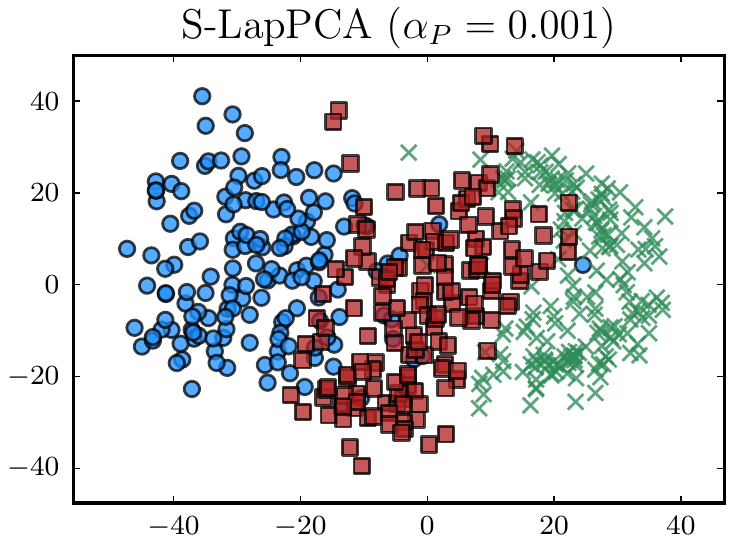}}
\subfloat[]{\includegraphics[width=0.19\linewidth]{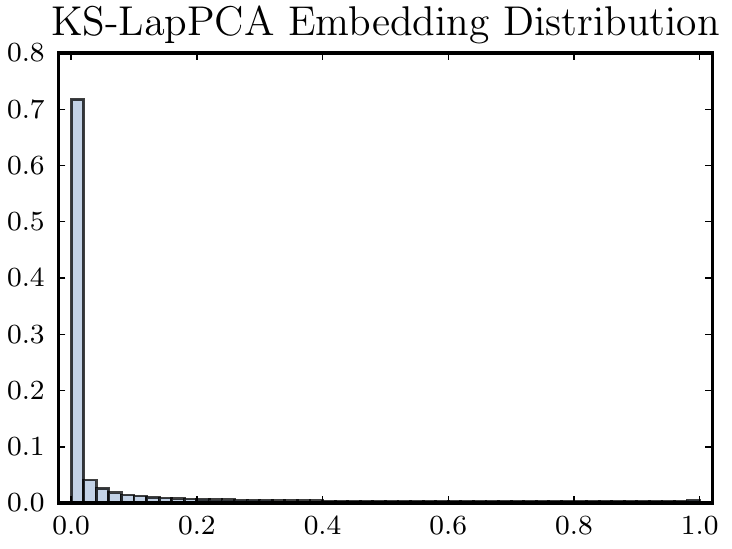}}
\subfloat[]{\includegraphics[width=0.19\linewidth]{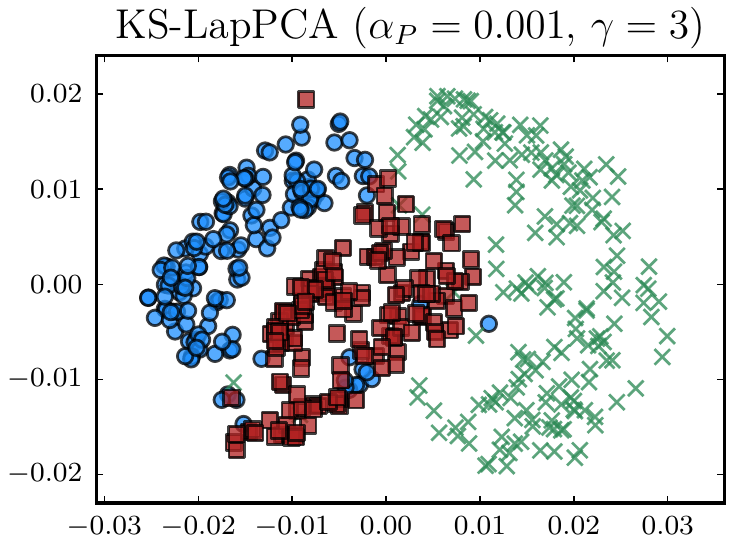}}

\subfloat[]{\includegraphics[width=0.19\linewidth]{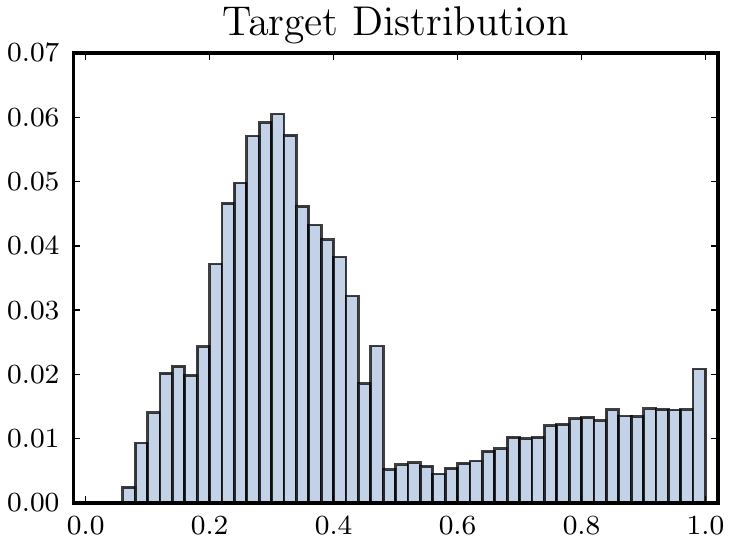}}
\subfloat[]{\includegraphics[width=0.19\linewidth]{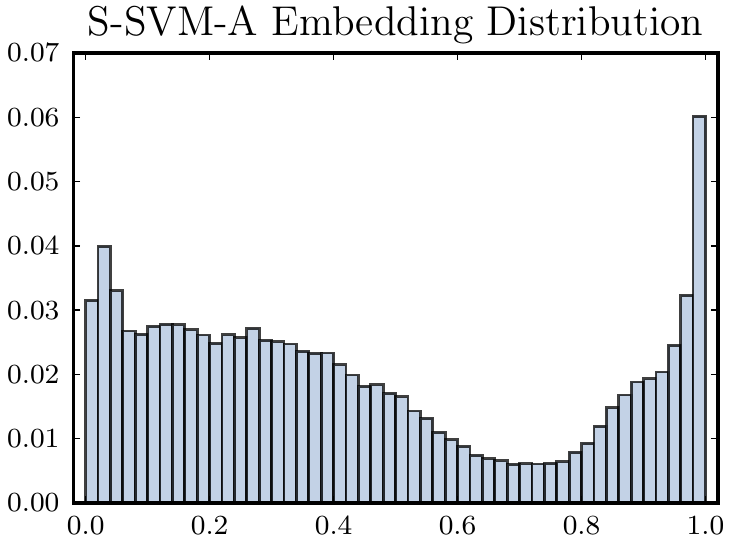}}
\subfloat[]{\includegraphics[width=0.19\linewidth]{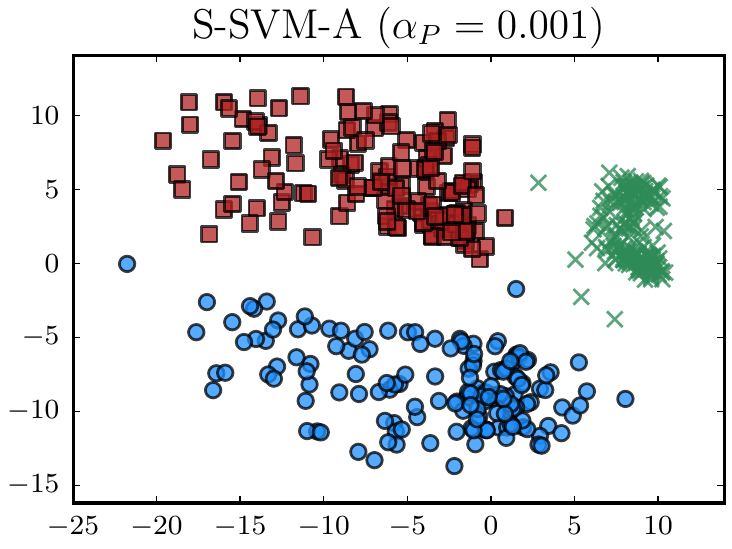}}
\subfloat[]{\includegraphics[width=0.19\linewidth]{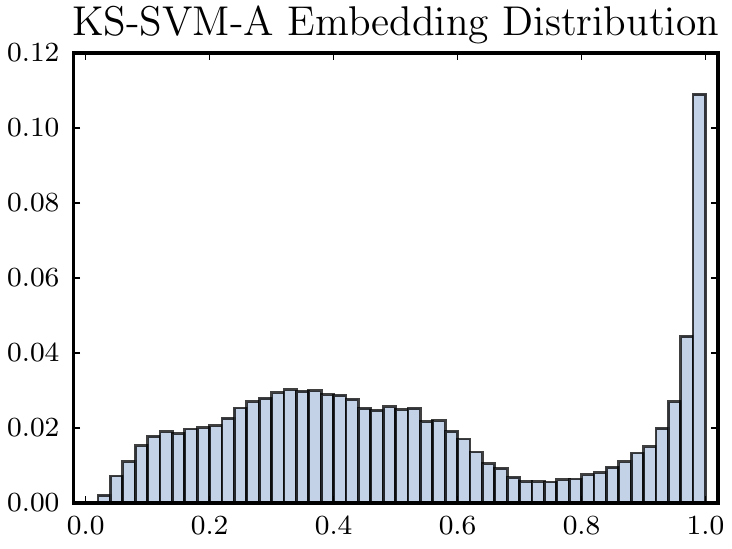}}
\subfloat[]{\includegraphics[width=0.19\linewidth]{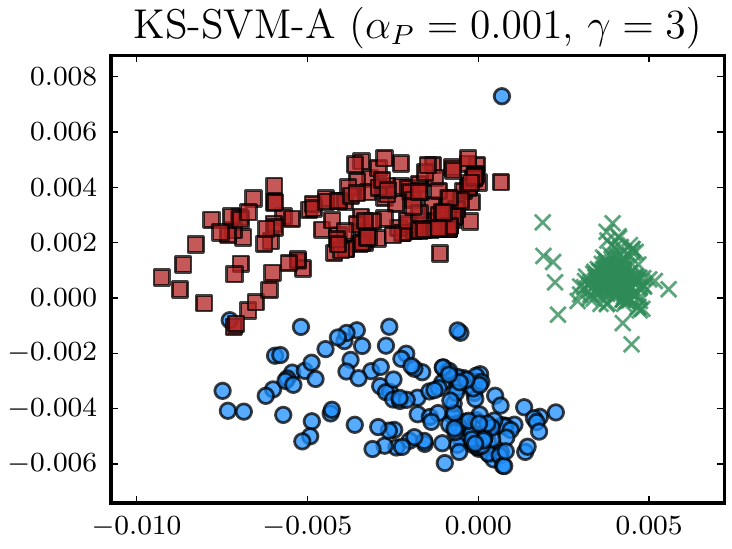}}
\end{center}

\caption{Using the SEF to perform various types of DR}
\footnotesize{Each row shows the results of a different DR method ((K)S-PCA, (K)S-LDA, (K)S-LapPCA, (K)S-SVM-A). The first column illustrates the target similarity distribution, the second and the third columns the achieved similarity distribution and the correponding low dimensional space using a linear projection function, while the other two columns repeats the previous two using a kernel projection function.}
\label{fig:toy-sef}
\end{figure*}

\paragraph{Principal Component Analysis} 

The Principal Component Analysis (PCA) \cite{pca}, is one of the most widely used techniques for dimensionality reduction. Although PCA is a linear technique, a kernel extension of PCA has been also proposed \cite{kpca}.  PCA seeks a projection that maximize the variance of the data in the projected space, i.e., maximizes the following criterion:
\begin{equation*}
J_{pca} = Var[\mathbf{Y'}] = \frac{1}{N} \sum_{i=1}^{N} ({y'}_i - E[\mathbf{Y'}])^2
\end{equation*} 
where $y'_i$ is a point in the projected space and $E[\mathbf{Y'}] =  \frac{1}{N} \sum_{i=1}^{N} {y'}_i$ is the mean of the projected points $\mathbf{Y'}$.

It can be easily derived that maximizing $J_{pca}$ is equivalent to maximizing the sum of the pairwise distances between the projected points \cite{kyperountas2010salient}. Also, maximizing the sum of the pairwise distances between the projected points is equivalent to minimizing the sum of the pairwise similarities. Therefore, to perform PCA using the proposed method the target similarity matrix is set to zero, i.e.,  $\mathbf{T}=\mathbf{T}_{PCA}=\mathbf{0}$. When this target similarity matrix is used, the SEF seeks embeddings that maximize the variance of the data in the low-dimensional space. Since this objective function is minimized by repeating the first principal component of the data $m$ times, it is important to set the orthonormality regularizer to a high value, e.g., $\alpha_p=1$.

In Figures \ref{fig:toy-sef}a-e the SEF is used to learn a PCA-like embedding. The distribution of the similarity values of the target similarity matrix $T$ (Figure  \ref{fig:toy-sef}a) and the distribution of the similarities of the projection similarity matrix (after the optimization) (Figures  \ref{fig:toy-sef}b/d) are also illustrated. After the projection, most of the similarity values are gathered near zero and the projected data are similar to those of a PCA projection. For the KS-PCA method, almost all the similarity values are within the range of the lower bucket of the histogram illustrated in Figure \ref{fig:toy-sef}d.

\paragraph{Linear Discriminant Analysis}

The Linear Discriminant Analysis (LDA) \cite{lda}, and its kernel extension \cite{kda-1}, is another well known technique for supervised dimensionality reduction. The LDA uses (an approximation to) the following optimization objective to learn a projection vector $\mathbf{w}$:
\begin{equation*}
J_{LDA} = \frac{\mathbf{w}^T \mathbf{S}_W \mathbf{w}}{\mathbf{w}^T \mathbf{S}_B \mathbf{w}} 
\end{equation*}
where $\mathbf{S}_W = \sum_{i=1}^{N} ( \mathbf{x}_i - E_i[\mathbf{X}]) ( \mathbf{x}_i - E_i[\mathbf{X}])^T$ is the intra-class scatter matrix, $\mathbf{S}_B = \sum_{i=1}^{N}  |\mathit{C}_i|(E_i[\mathbf{X}] - E[\mathbf{X}])(E_i[\mathbf{X}] - E[\mathbf{X}])^T$ is the inter-class scatter matrix, $\mathit{C}_i$ is the set that contains the points that belong to the $i$-class and $E_i[\mathbf{X}]$ is the mean vector of the points that belongs to $\mathit{C}_i$.

The intra-class scatter is minimized when the sum of the pairwise distances between the points that belong to the same class is minimized, while the inter-class scatter is maximized when the sum of the pairwise distances between the points that belong to different classes is maximized. Therefore, to perform LDA using the proposed technique the target matrix is set as:
\begin{equation*}
[\mathbf{T}]_{ij} =
\begin{cases}
1, \text{if the points $i$ and $j$ belong to the same class}\\
0, \text{otherwise}
\end{cases}
\end{equation*}
where the target similarity of the points that belong to the same class is set to 1, and the target similarity of the points that belong to different classes is set to 0. To ensure that minimizing the intra-class scatter and maximizing the inter-class scatter is equally important the optimization mask $\mathbf{M}$ is set as:
\begin{equation}
\label{eq:mask}
[\mathbf{M}]_{ij} =
\begin{cases}
1, \text{if the points $i$ and $j$ belong to the same class}\\
\frac{1}{N_C-1}, \text{otherwise}
\end{cases}
\end{equation}
where $N_C$ is the total number of classes. Note that the LDA method is limited to learning a maximum of $N_C-1$ projection directions. This upper limit does not affect the proposed (K)S-LDA. Thus, any number of projection directions can be used, possibly increasing the accuracy of the used classification algorithm.

The S-LDA and the KS-LDA methods are illustrated in Figure \ref{fig:toy-sef}f-j. The KS-LDA  method manages to almost collapse the classes into three distinct points in the learned space, while the S-LDA to clearly separate the three different classes. Also, the effect of the target similarity distribution on the actual projection distribution is evident in the Figures \ref{fig:toy-sef}g/i.

\paragraph{Laplacian Eigenmaps}
The Laplacian Eigenmaps (LE) method \cite{belkin2001laplacian}, along with its linear variant called Locality Preserving Projection (LPP),  try to preserve the local relations between neighboring points by minimizing the following objective:

\begin{dmath*}
J_{Laplacian} = \sum_{i=1}^N \sum_{j=1}^N [\mathbf{W}_L]_{ij}({y'}_i - {y'}_j)^2
\end{dmath*}
where the $\mathbf{W}_L$ is the adjacency matrix of the data. Several ways exist to define the adjacency matrix $\mathbf{W}_L$. When the adjacency matrix is defined as:

\begin{dmath*}
[\mathbf{W}_L]_{ij} =
\begin{cases}
1, \text{if $i$ is among the $k$ nearest neighbors of $j$}\\
0, \text{otherwise}
\end{cases}
\end{dmath*},
the objective is minimized when the distance between the neighboring points is minimized.
To obtain LE-like embeddings using the proposed technique the target similarity matrix is set to:
\begin{dmath*}
[\mathbf{T}]_{ij} =
\begin{cases}
1, \text{if $i$ is among the $k$-nearest neighbors of $j$}\\
 exp(-\frac{||\mathbf{x}_i - \mathbf{x}_j||_2^2}{\sigma_{LE}}), \text{otherwise}
\end{cases}
\end{dmath*}

Note that for the non-neighboring points the similarity target is set to the original similarity of the corresponding data points (using the scaling factor $\sigma_{LE}$), while for neighboring points to 1. Although, it is not necessary to constraint the similarity of the non-neighboring points, this ensures that degenerate solutions, such as collapsing all the points into one point, will be avoided. The weight of the constraint for the non-neighboring points can be set be defining an appropriate optimization mask (similarly to (\ref{eq:mask})).

\paragraph{Laplacian PCA}
In the previous technique the similarity of the non-neighboring points was set to the original similarity between the data points. Another options is to request non-neighboring points to be as far apart as possibly by setting their similarity to 0. That is, the proposed method seeks projections that minimizes the distance between the neighboring points (Laplacian-like objective), while maximizing the variance of the rest data points (\mbox{PCA-like} objective). Using this approach, a method similar to the Laplacian~PCA (abbreviated as LapPCA) \cite{laplacian-pca}, is derived. An example of learning Laplacian \mbox{PCA-like} embeddings using the \mbox{S-LapPCA} and \mbox{KS-LapPCA} techniques is shown in Figures \ref{fig:toy-sef}k-o (2000 optimization iterations are used for the kernel method). The figure of the kernel embedding (Figure \ref{fig:toy-sef}o) clearly illustrates the gathering of the neighboring points and the separation of non-neighboring ones. Note that semi-supervised methods \cite{belkin2001laplacian, huang2012semi}, can be similarly derived (by appropriately setting 0 and 1 for the data pairs with known labels).

\paragraph{t-distributed stochastic neighbor embedding}

The \mbox{t-distributed} stochastic neighbor embedding (t-SNE) algorithm \cite{van2008visualizing}, is a non-parametric technique for dimensionality reduction that is well suited for visualization, since it solves the crowding problem that can occur when high-dimensional data are embedded in low-dimensional spaces with few dimensions.  The t-SNE algorithm can be considered as a special case of the proposed framework where the similarity in the projected space is calculated using a heavy-tailed \mbox{Student-t} distribution: $
[\mathbf{P}]_{ij} = \frac{(1+ ||\mathbf{y}_i - \mathbf{y}_j||_2^2)^{-1}}{\sum_{k=1, k\neq i} (1+ ||\mathbf{y}_i - \mathbf{y}_k||_2^2)^{-1}}
$. The target similarity matrix is calculated as $[\mathbf{T}]_{ij} = \frac{p_{ij}+p_{ji}}{2N}$, where $
p_{ij} =\frac{exp(-||\mathbf{x}_i-\mathbf{x}_j||^2/\sigma_{t-SNE}^2)}{\sum_{k=1, k \neq i}^N exp(-||\mathbf{x}_i-\mathbf{x}_k||^2/\sigma_{t-SNE}^2)}$. Finally, instead of using the weighted squared loss as the objective, the Kullback-Leibler divergence is used: $
J_{t-SNE} = \sum_{i=1}^{N} \sum_{J=1}^{N} [\mathbf{T}]_{ij} \log \frac{ [\mathbf{T}]_{ij}}{ [\mathbf{P}]_{ij}}$.

\subsubsection{Cloning An Existing Technique}
\label{section:out-of-sample}

\begin{figure}
{\includegraphics[width=0.32\linewidth]{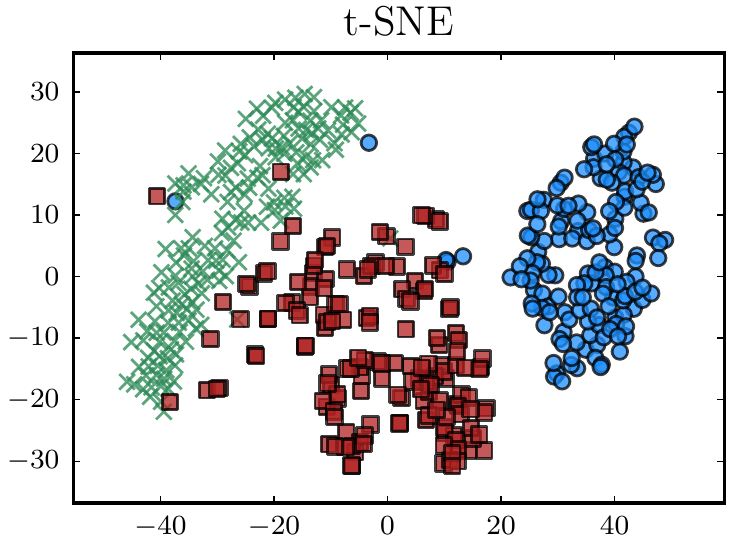}}
{\includegraphics[width=0.32\linewidth]{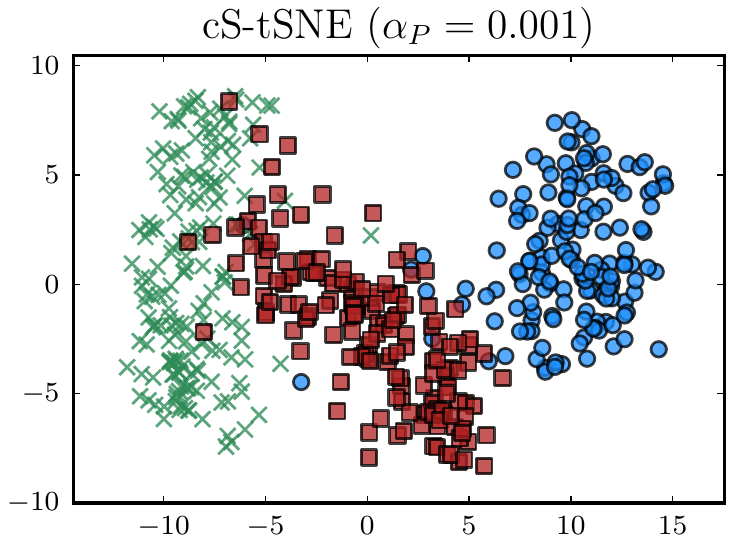}}
{\includegraphics[width=0.32\linewidth]{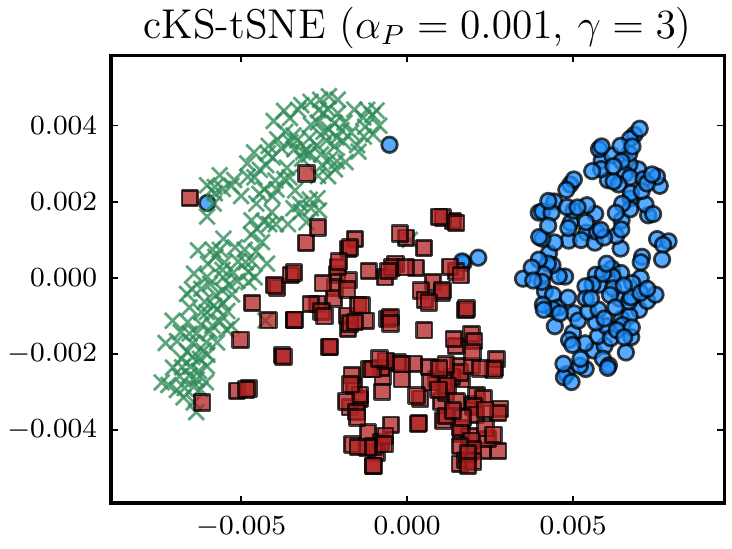}}\\
{\includegraphics[width=0.32\linewidth]{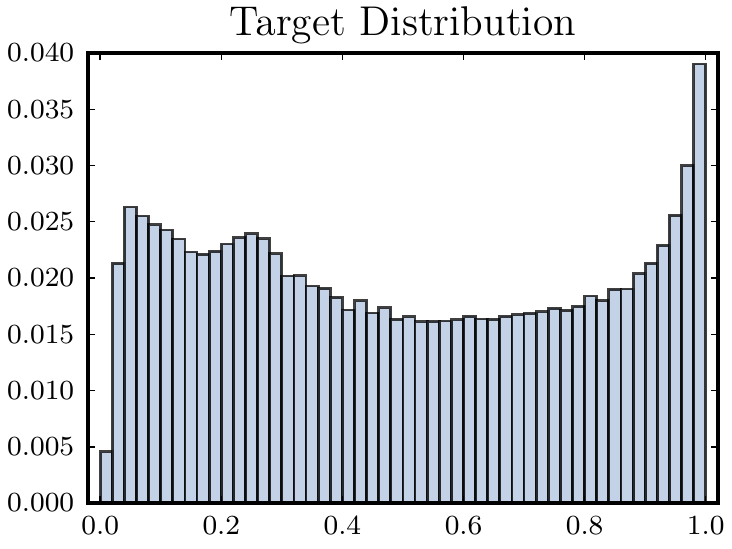}}
{\includegraphics[width=0.32\linewidth]{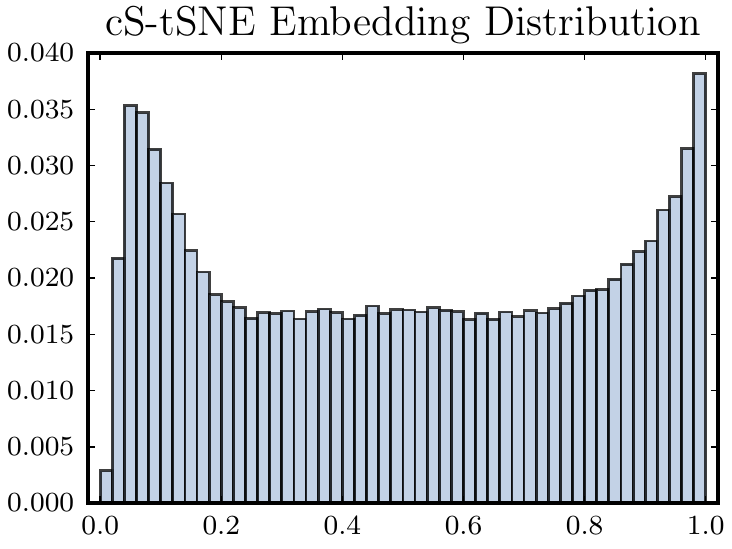}}
{\includegraphics[width=0.32\linewidth]{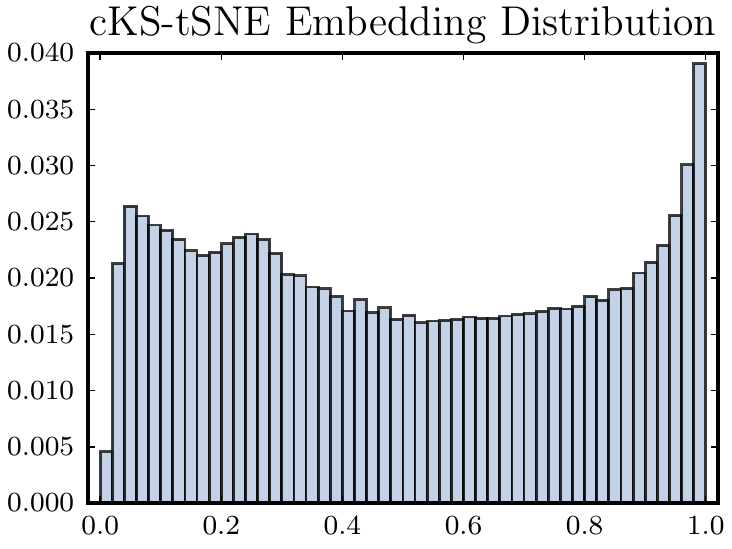}}

\caption{Using the cS-tSNE and the cKS-tSNE methods to ``mimic'' another method  (t-SNE)}
\label{fig:toy-copy}
\end{figure}

The proposed framework can be also used to ``clone'' an existing technique by learning an embedding that approximates the distribution of the points that the method to be cloned provides. Let $g(\mathbf{x})$ be an arbitrary technique that provides an embedding for the points $\mathbf{x} \in \mathcal{X}_{train}$. The SEF can learn an embedding that mimics  the method $g(\mathbf{x})$ by defining the target matrix as:

\begin{equation}
\label{eq:out-of-sample-similarity}
[\mathbf{T}]_{ij} = exp(-\frac{||g(\mathbf{x}_i) - g(\mathbf{x}_j)||_2^2}{\sigma_{copy}})
\end{equation}
Again, the parameter $\sigma_{copy}$ is chosen in such way to maximize the spread of the similarity values. 

The method learns an embedding that ``copies'' the target space into the projected space. Note that the SEF does not perform regression on the actual representation of the data. Instead it learns a projection that approximates the similarity between every pair of the embedded points. This allows for using an arbitrary number of dimensions on the projected space, which is not possible when regression techniques are used. That also means that the representation of the points in the learned space will be different from the target representation and the training set of data must be also projected to the learned space. The linear embedding is denoted by \textit{cS-XYZ}, while the kernel embedding by \textit{cKS-XYZ}, where \textit{XYZ} is the name of the technique to be cloned.

The most straightforward way to use the cloning ability of the SEF is to provide out-of-sample extensions. This is illustrated in Figure \ref{fig:toy-copy}, where the t-SNE algorithm is used as the target technique. The target similarity distribution and the similarity distribution of the embeddings are also shown in the second row of Figure \ref{fig:toy-copy}. The kernel technique manages to accurately recreate the target space (2000 optimization iterations were used). 

This technique can be also used to linearly (or non-linearly) approximate more complex techniques, such as  kernel methods or techniques that work on higher dimensional spaces. The latter is possible since the SEF does not perform regression on the actual representation of the data. In Section \ref{section:experiments-fast} it is demonstrated that this method can significantly increase the accuracy of PCA, while using less dimensions.

\subsubsection{SVM-based Projection}
\label{section:svm-based}

New DR techniques can be also derived within the proposed framework by simply manipulating the target similarity matrix, i.e., the target PDF.  In this Subsection, a novel analysis technique, based on the Support Vector Machine classification model \cite{james2013introduction}, is proposed. The SVMs classify the data into two categories by calculating the hyperplane that separates the data by the largest margin. The SVMs are binary classifiers, i.e., they can only classify the data into two categories. Although multi-class SVM formulations exist \cite{crammer2002algorithmic}, usually the one-versus-one and the one-versus-all methods are used to handle multi-class data \cite{james2013introduction}. In the one-versus-one technique a binary classifier is trained for each pair of two classes, requiring $\frac{N_C (N_C-1)} {2}$ separate models to classify the data. Alternative, the one-versus-all technique can be used, requiring $N_C$ separate models. In both cases, a significant overhead is added, since every new sample must be fed to each model and then the individual decisions must be aggregated.

In the SVM-based embedding the target similarity matrix is constructed in such way that the obtained embedding will approximate the relations between the points as expressed by the learned SVM models. Each SVM defines a one-dimensional representation for the points that separates, which is given by the (signed) distance of each point to the separating hyperplane (multiplied by 1 or -1 according to the side of the hyperplane that each point lies). Therefore, the target similarity matrix is defined as: $
\mathbf{T} = exp(-\frac{\mathbf{D}_{SVM}}{\sigma_{SVM}})$, where 
\begin{dmath*}
[\mathbf{D_{SVM}}]_{ij} = abs(SVM_{(c(i), c(j))}(\mathbf{x}_i ) - SVM_{(c(i), c(j))}(\mathbf{x}_j ))
\end{dmath*}
is a pairwise distance matrix. The notation $c(i)$ is used to refer to the class of the sample $\mathbf{x}_i$ and $SVM_{(c(i), c(j))}(\mathbf{x}_i )$ is the representation of $\mathbf{x}_i$ (signed distance to the hyperplane) using the SVM model that separates the classes of the $i$-th and the $j$-th points. The parameter $\sigma_{SVM}$ is chosen using the same heuristic line-search method used for $\sigma_P$.

After projecting the data using the SVM-based target similarity matrix, a light-weight classifier, such as the Nearest Centroid Classifier (NCC), can be used to provide a fast classification decision. The following process can be then followed to quickly classify a new data sample: a) the data sample is projected to the learned low-dimensional space, and b) it is assigned to the class centroid that is closer to its projection. This avoids the time-consuming step of feeding each new point to a separate classifier and then aggregating the resulting decisions. Note that the proposed technique can work with any type of SVM (linear or kernel). The speed benefits are expected to be larger for kernel SVMs.

A toy example of the SVM-based projection is provided in Figures \ref{fig:toy-sef}p-t. The proposed technique is called S-SVM-A(nalysis) (KS-SVM-A for the kernel technique). Both the linear and the kernel embedding manages to decently separate the three classes that exist in the data. However, in contrast to the LDA technique, which collapses the points when a kernel projection is used (Figures \ref{fig:toy-sef}g/j), the SVM-based projection avoids this phenomenon. This can reduce the risk of overfitting the learned projections. On the other hand, the existence of support vectors in the wrong side of the separating hyperplane seems to produce a few outliers. Nonetheless, it is experimentally demonstrated in the subsequent Section that the proposed technique works quite well in practice. Finally, note that instead of utilizing a SVM, any classifier that can (directly or indirectly) estimate the probability of two points belonging to the same class can be used to derive similar analysis-based techniques.

\section{Experiments}
\label{section:experiments}

In this Section, the experimental evaluation of the proposed framework is provided. First, the used datasets and the evaluation protocol are described. Then, the SEF is evaluated using 4 different setups: a) providing a linear low-dimensional embedding for a high dimensional DR technique b) using the S-LDA supervised embeddings, c) using the proposed SVM-based analysis technique and d) providing (linear and kernel) out-of-sample extensions for a manifold-based technique. 

\subsection{Datasets and Evaluation Protocol}
For evaluating the proposed techniques, the following six datasets from a wide range of domains are used: two multi-class image recognition datasets, the 15-scene dataset \cite{lazebnik2006beyond}, and the Corel dataset \cite{corel-dataset}, one multi-class video dataset, the KTH action recognition database \cite{schuldt2004recognizing}, one hand-written digit recognition dataset, the MNIST dataset \cite{lecun1998mnist}, one face recognition dataset, the extended Yale B dataset (abbreviated as Yale) \cite{ KCLee05}, and one text dataset for topic recognition, the 20 Newsgroups dataset (abbreviated as 20NG) \cite{ng20}. 

The 15-scene dataset \cite{lazebnik2006beyond}, contains 15 different scene categories. The total number of images is 4,485 and each category has 200 to 400 images. HoG \cite{dalal2005histograms}, and LBP features \cite{ojala2002multiresolution}, of 8 $\times$ 8 non-overlapping patches are densely extracted from each image. The two feature vectors extracted from each patch are fused together to form the final feature vector. Then, the BoF model \cite{sivic2003video}, is used to learn a dictionary (using the k-means algorithm) and extract a 512-dimensional histogram vector for each image.  The Corel dataset \cite{corel-dataset}, contains 10,800 images from 80 different concepts. Again, HoG and LBP features of 8 $\times$ 8 patches are densely extracted from each image and histogram vectors are compiled for the images using the BoF model.

The MNIST database \cite{lecun1998mnist}, is a well-known dataset that contains 60,000 training and 10,000 testing images of handwritten digits. There are 10 different classes, one for each digit (0 to 9), while the size of each image is 28 $\times$ 28. Each image is represented using its pixel representation, after flatting it to a 784-dimensional vector.

The cropped Extended Yale Face Database B \cite{KCLee05}, contains 2432 images from 38 individuals. The images were taken under greatly varying lighting conditions. The size of each image is $168 \times 192$ and, similarity to the MNIST dataset, the raw pixel representation is used for each image.

The KTH action recognition dataset \cite{schuldt2004recognizing},  contains 2391 video sequences of six types of human actions (walking, jogging, running, boxing, hand waving and hand clapping). The train+validation (1528 videos) and the test (863 videos) splits are predefined. From each video, HoG and HoF descriptors are extracted \cite{laptev2008learning}. For each type of descriptor a BoF dictionary with 512 codewords is learned. Then, each video is represented by fusing the extracted histogram vectors.

The 20 Newsgroups dataset  \cite{ng20}, contains 18,846  documents that belong to 20 different newsgroups. The train (11,314 documents) and the test split (7,532 documents) are predefined. The popular tf-idf method \cite{manning-etal}, is used to represent the documents. The dictionary is pruned by discarding the top 5\% most frequent words and the 1\% of the less frequent words, leading to a textual dictionary with 2164 words.

For the 15-scene 100 images are randomly sampled from each class to form the training set. The rest of the images are used to test the learned model. For the Corel dataset the training set is composed of 60 randomly sampled images from each class, while for the Yale dataset, 30 images are sampled for each person. Again, the rest of the images are used to test the evaluated method. For the datasets that have predefined train/test splits, i.e., the MNIST, the KTH dataset and the 20NG dataset, at most 5000 randomly chosen samples are used to train the models. The whole test set is always used for the evaluation. The evaluation process is repeated 10 times and the mean and the standard deviation of the evaluation criteria are reported. For the KTH dataset, the same (shuffled) training set is used (since less than 5000 training samples exist). However, the hyper-parameter  selection procedure for the SVM classifier uses 3-fold cross validation, which can lead to non-zero standard deviation  for some experiments.

For training the linear methods, 500 optimization iterations are used, while for training the kernel methods 1000 optimization iterations are used. The method is generally stable to the value used for the orthonormality regularizer weight $\alpha_p$ and the optimal value seems to depend mostly on the type of the learning task. To choose the optimal value for the $\alpha_p$ parameter the training set was split into a new training and validation set. Although a different value of $\alpha_p$ is chosen for each type of experiment, the same value is utilized for all the datasets used for a specific type of experiment. The only exception to this is the supervised evaluation experiments, where for the Yale dataset a different value of $\alpha_p$ is used (as discussed later). For the cS-PCA, the S-LDA and the cS-ISOMAP $\alpha_p$ is set to 1, while for the cKS-ISOMAP and the S-SVM-A $\alpha_p$ is set to 0.001. 

\subsection{Approximating a high dimensional technique using less dimensions}
\label{section:experiments-fast}

\begin{table*}
	\caption{Approximating a high dimensional technique using less dimensions (SVM classification accuracy)}	
	\label{table:la-2}
	\centering
	\begin{tabular}{l|cc||cc}
		\textbf{dataset} &\textbf{PCA ($10$)} &\textbf{cS-PCA ($10$)} &\textbf{PCA} (5 samples, $10$) &\textbf{cS-PCA} (5 samples, $10$) \\ 
		\hline 
		\textbf{15-scene} &  $62.83 \pm 0.96$ &  $\mathbf{66.68 \pm 0.54}$ & $42.77 \pm 1.86$ &  $\mathbf{49.06 \pm 2.79}$ \\ 
		\textbf{Corel} &  $36.11 \pm 0.56$ &  $\mathbf{38.35 \pm 0.37}$ & $25.91 \pm 1.28$ &  $\mathbf{27.95 \pm 1.28}$\\ 
		\textbf{MNIST} &  $83.78 \pm 0.37$ &  $\mathbf{85.68 \pm 0.40}$ & $\mathbf{61.44 \pm 2.90}$ &  ${59.85 \pm 3.55}$\\ 
		\textbf{Yale} &  $56.79 \pm 1.80$ &  $\mathbf{63.81 \pm 1.93}$ & $35.98 \pm 3.75$ &  $\mathbf{38.93 \pm 2.95}$\\ 
		\textbf{KTH} &  $77.94 \pm 0.37$ &  $\mathbf{84.19 \pm 0.44}$ & $66.43 \pm 6.52$ &  $\mathbf{72.16 \pm 6.50}$\\ 
		\textbf{20NG} &  $40.46 \pm 0.82$ &  $\mathbf{46.74 \pm 0.90}$ & $14.96 \pm 1.60$ &  $\mathbf{14.99 \pm 1.26}$\\ 
	\end{tabular}
\end{table*}

\begin{table*}
	\caption{Supervised Embeddings (SVM classification accuracy)}	
	\label{table:sup-svm}
	\centering
	\begin{tabular}{l|ccc|cc}
		\textbf{dataset} &\textbf{LDA ($N_C-1$)} &\textbf{S-LDA ($N_C-1$)} &\textbf{S-LDA ($2 \times N_C-1$)}  & \textbf{LDA} (5 samples, $N_C-1$) &\textbf{S-LDA} (5 samples, $2 \times N_C-1$) \\ 
		\hline 
		\textbf{15-scene} &  $68.17 \pm 0.95$ &  $75.39 \pm 0.70$ &  $\mathbf{75.87 \pm 0.76}$ & $40.14 \pm 2.71$ & $\mathbf{52.09 \pm 2.85}$ \\ 
		\textbf{Corel} &  $37.75 \pm 0.49$ &  $\mathbf{44.00 \pm 0.43}$ &  $43.60 \pm 0.41$ & $17.26 \pm 0.79$ & $\mathbf{28.27 \pm 0.66}$\\ 
		\textbf{MNIST} &  $86.10 \pm 0.22$ &  $89.17 \pm 0.31$ &  $\mathbf{89.53 \pm 0.26}$ & $49.01 \pm 3.78$ & $\mathbf{60.48 \pm 3.11}$\\ 
		\textbf{Yale} &  $\mathbf{95.21 \pm 0.83}$ &  $93.59 \pm 0.51$ &  $94.37 \pm 0.70$ & $59.36 \pm 3.46$ & $\mathbf{71.87 \pm 1.49}$\\ 
		\textbf{KTH} &  $90.38 \pm 0.00$ &  $87.07 \pm 1.22$ &  $\mathbf{92.58 \pm 0.00}$ & $65.55 \pm 2.79$ & $\mathbf{76.83 \pm 4.79}$\\ 
		\textbf{20NG} &  $63.66 \pm 0.52$ &  $\mathbf{69.97 \pm 0.30}$ &  $69.54 \pm 0.39$ & $9.68 \pm 1.35$ & $\mathbf{20.03 \pm 2.70}$\\ 
	\end{tabular}
\end{table*}

\begin{figure}
\includegraphics[width=0.31\linewidth]{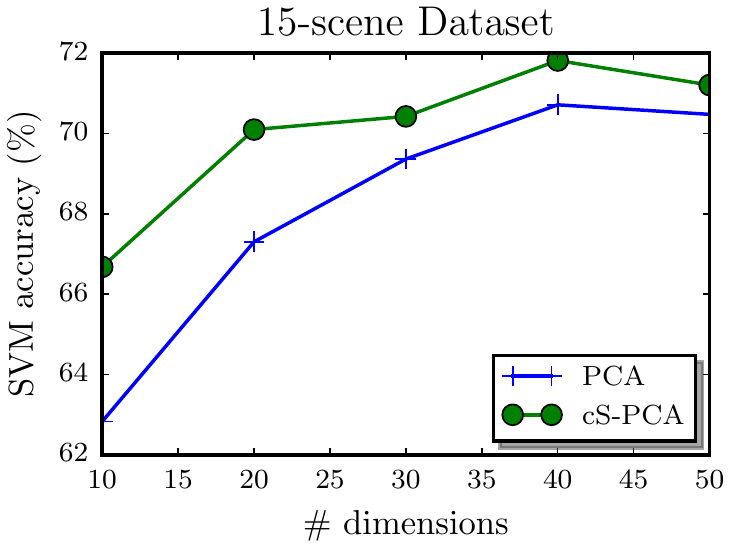}
\includegraphics[width=0.31\linewidth]{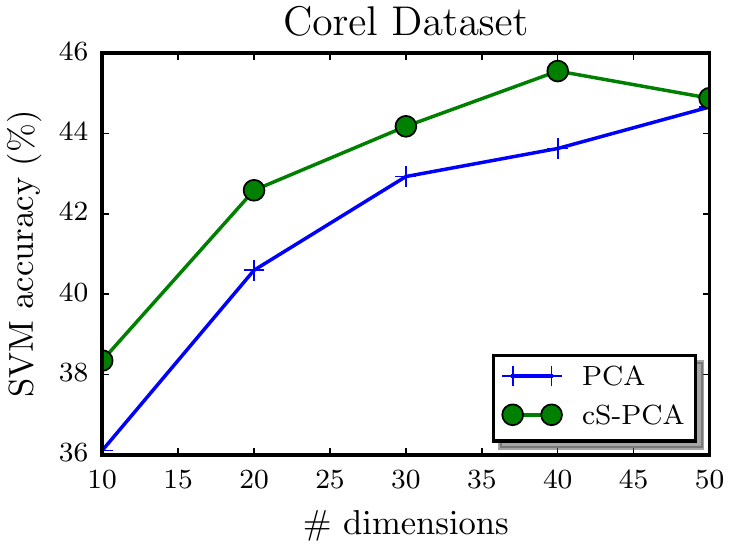}
\includegraphics[width=0.31\linewidth]{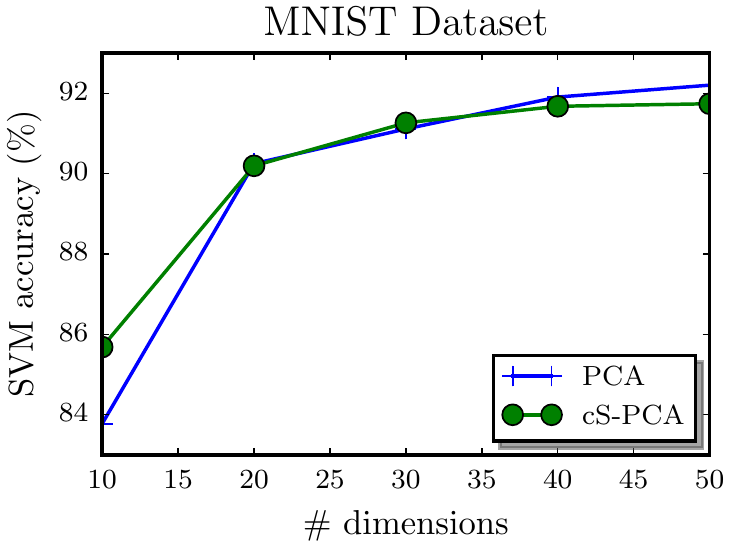}\\
\includegraphics[width=0.31\linewidth]{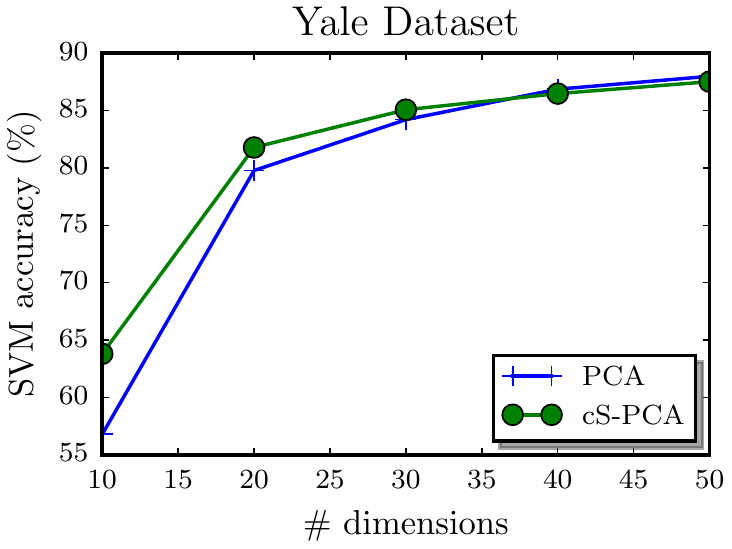}
\includegraphics[width=0.31\linewidth]{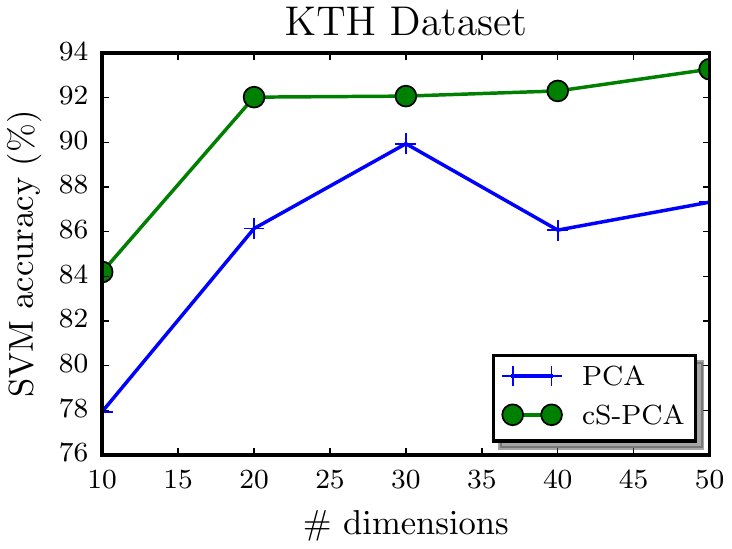}
\includegraphics[width=0.31\linewidth]{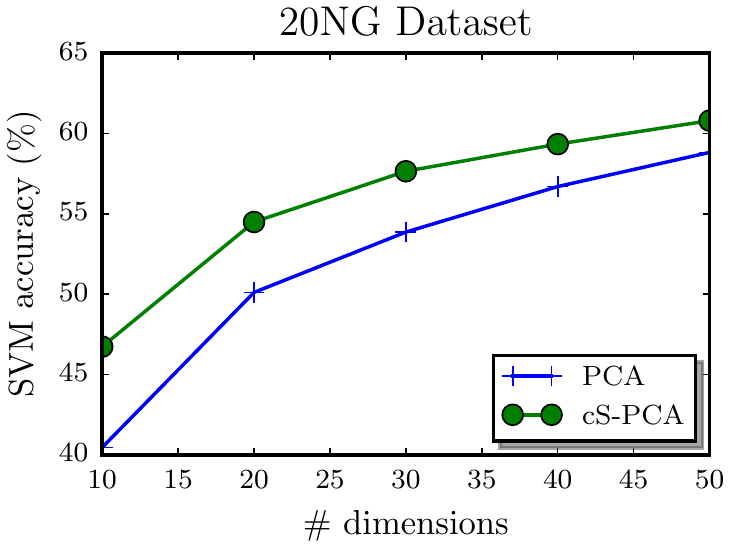}
\caption{The accuracy of the PCA and cS-PCA methods are plotted for different number of dimensions. The target for the cS-PCA method is the 50-dimensional PCA.}
\label{fig:la-1}
\end{figure}

First, the cS-PCA method is used to approximate the 50-dimensional PCA using only 10 dimensions. To this end, the similarity matrix of the 50-dimensional PCA is used as the target similarity matrix, by utilizing the technique described in Subsection \ref{section:out-of-sample}. Note that the learned projection is not expected to obtain better solutions in terms of the objective function of the used technique, i.e., achieve greater variance in the projected space than the PCA method. The reasoning behind this method is that a more complex or more high dimensional technique better separates the data. Therefore, part of the high-dimensional structure of the data can be embedded in a lower-dimensional space, leading to better classification accuracy.

In Table \ref{table:la-2} the cS-PCA method is compared to the PCA method using either the full training data (left part of Table~\ref{table:la-2}) or only 5 training samples per class (right part of Table~\ref{table:la-2}). The dimensionality is set to $m=10$ for both methods. A linear SVM is used to evaluate the learned representations. For selecting the parameter $C$ of the SVM 3-fold cross validation is used. The cS-PCA almost always performs better than the PCA technique, even when the training data are limited to just 5 samples per class. Extensive experiments using varying number of training samples per class are included in the Supplementary Material (Section 2).   Furthermore, the paired t-test was used to validated the statistical significance of the obtained results \cite{gibbons2011nonparametric}. The null
hypothesis (`There is no statistically significant difference between the cS-PCA~(10) and the PCA~(10) methods.') is rejected ($p = 0.004 < a = 0.05$).

In Figure \ref{fig:la-1} the accuracy of PCA and cS-PCA methods is plotted for different number of dimensions and datasets. The target for the cS-PCA method is always the 50-dimensional PCA. Again, the cS-PCA almost always performs better than the same-dimensional PCA technique. As the dimensionality of the learned space increases the two methods converge (which is expected), except for the KTH dataset where the cS-PCA  performs much better than the method that it clones. This behavior can be explained by considering that the cS-PCA method is intrinsically resistant to producing outliers due to the non-linear transformation of the pairwise distances. Therefore, it can be also seen as a more-regularized version of the PCA method, which can improve the classification accuracy. 

\begin{table*}
\caption{SVM-based Analysis (Classification accuracy)}
\label{table:svm}
\begin{center}
\begin{tabular}{l|ccc||c}
\textbf{dataset} &\textbf{Raw ($D$) + NCC} &\textbf{ S-SVM-A ($N_C$) + NCC } &\textbf{S-SVM-A  ($2 \times N_C$) + NCC } & \textbf{Raw ($D$) + SVM} \\ 
\hline 
\textbf{15-scene} &  $59.20 \pm 1.21$ (3.32 $\mu$sec)  &  $\mathbf{73.05 \pm 0.80}$ (1.83 $\mu$sec) &  $72.66 \pm 0.68$ (1.99 $\mu$sec) &  $74.23 \pm 0.42$ (595.47 $\mu$sec) \\ 
\textbf{Corel} &  $37.55 \pm 0.48$  (5.24 $\mu$sec)&  $41.67 \pm 0.47$ (3.14 $\mu$sec)&  $\mathbf{42.44 \pm 0.68}$ (3.89 $\mu$sec) &  $47.06 \pm 0.56$ (2567.58 $\mu$sec)\\ 
\textbf{MNIST} &  $81.64 \pm 0.34$ (4.12 $\mu$sec) &  $\mathbf{87.28 \pm 0.34}$ (2.43 $\mu$sec) &  $87.27 \pm 0.13$ (2.50 $\mu$sec)  &  $92.46 \pm 0.15$ (1395.44 $\mu$sec)\\ 
\textbf{Yale} &  $12.38 \pm 1.38$ (160.12 $\mu$sec) &  $86.95 \pm 0.79$ (102.35 $\mu$sec) &  $\mathbf{90.11 \pm 0.96}$ (106.94 $\mu$sec) &  $91.75 \pm 0.94$  (30432.12 $\mu$sec) \\ 
\textbf{KTH} &  $79.72 \pm 0.00$ (5.19 $\mu$sec) &  $\mathbf{93.16 \pm 0.00}$ (3.63 $\mu$sec) &  $92.70 \pm 0.00$  (3.65 $\mu$sec) &  $89.15 \pm 0.23$ (168.87 $\mu$sec) \\ 
\textbf{20NG} &  $60.40 \pm 0.45$ (9.10 $\mu$sec) &  $62.24 \pm 0.46$ (15.99 $\mu$sec) &  $\mathbf{63.78 \pm 0.41}$ (16.58 $\mu$sec) &  $68.37 \pm 0.75$ (8051.48 $\mu$sec)\\ 
\end{tabular}\\
\end{center}
(The quantities in the parenthesis refer to the mean classification time \textit{per sample} (total testing time divided by the number of testing samples))
\end{table*}

\subsection{Supervised Embeddings}

\begin{figure}
\includegraphics[width=0.32\linewidth]{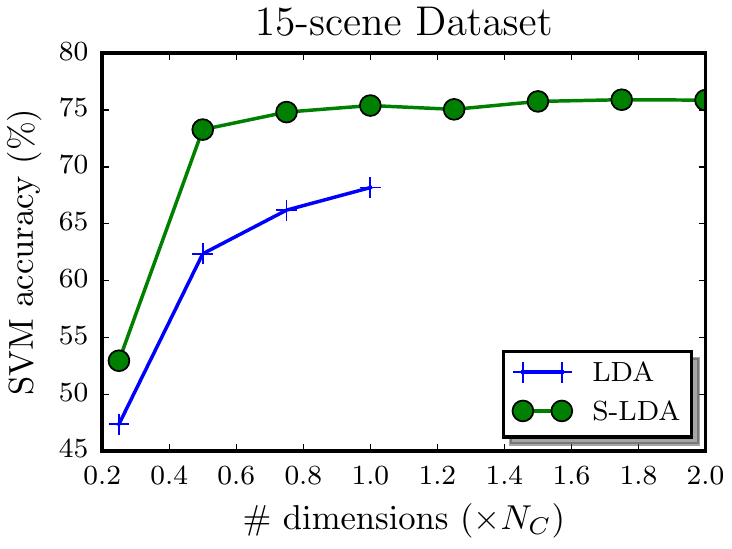}
\includegraphics[width=0.32\linewidth]{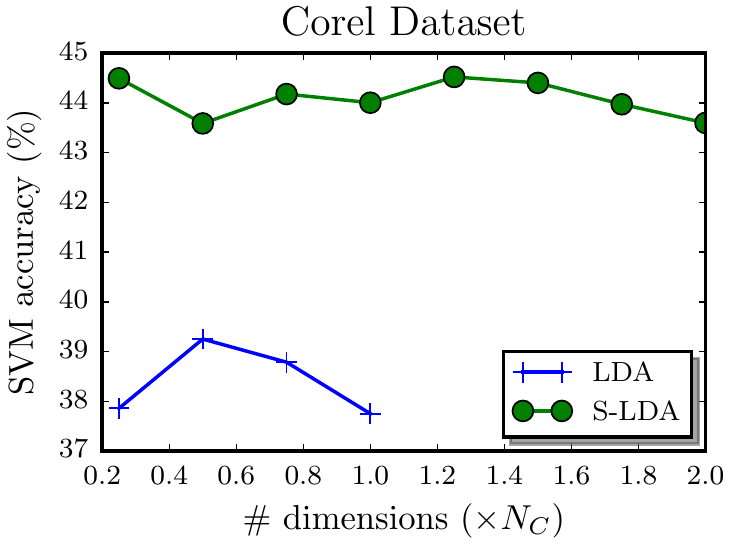}
\includegraphics[width=0.32\linewidth]{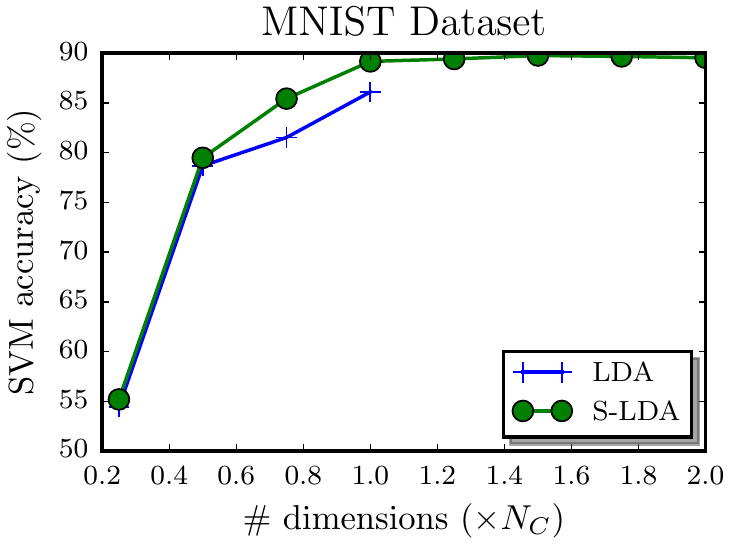} \\
\includegraphics[width=0.32\linewidth]{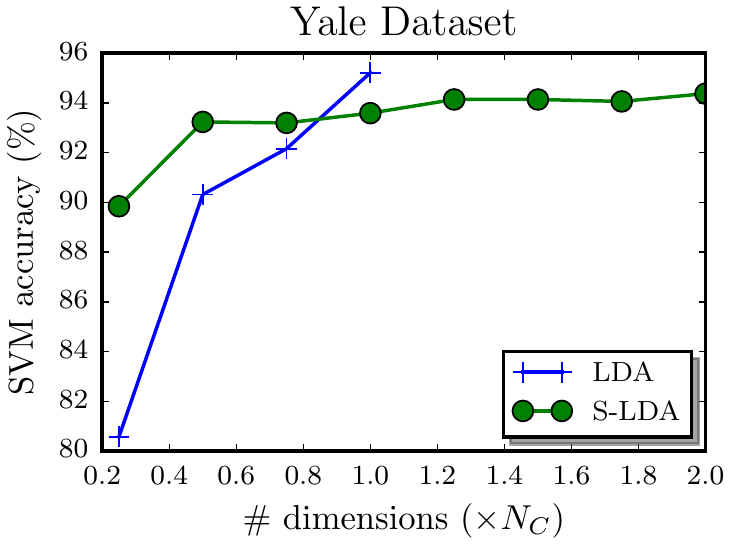}
\includegraphics[width=0.32\linewidth]{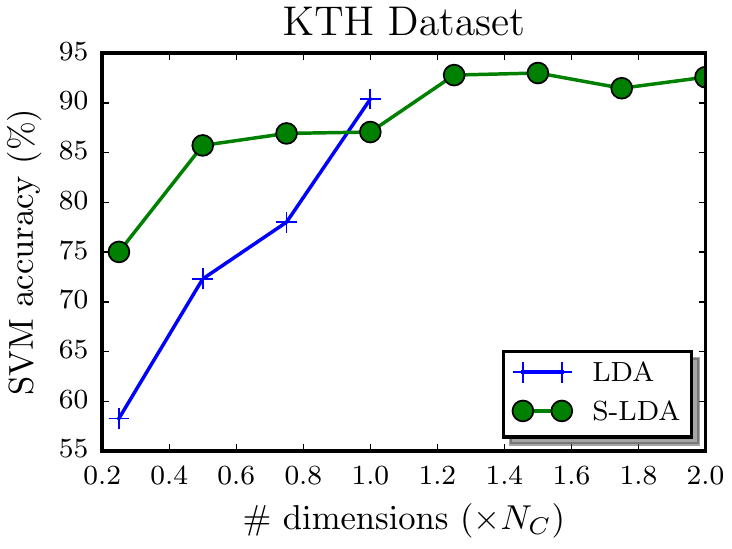}
\includegraphics[width=0.32\linewidth]{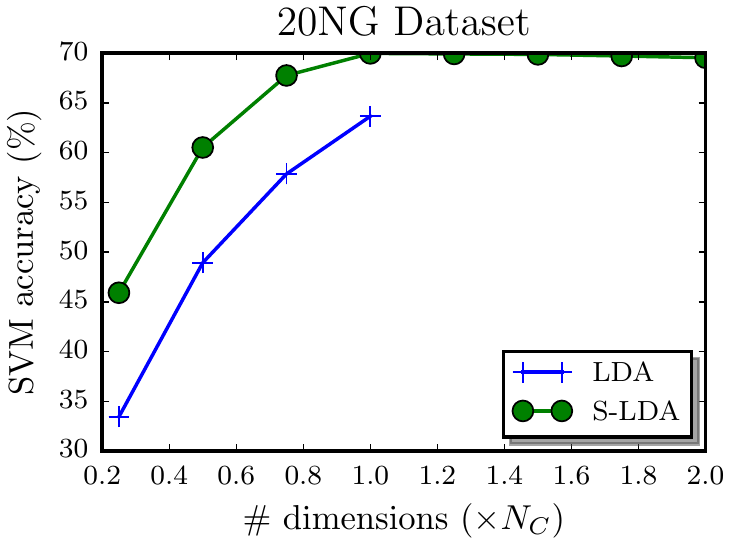}
\caption{The accuracy of the LDA and cS-LDA methods are plotted for different number of dimensions.}
\label{fig:sup-svm-curves}
\end{figure}

The SEF can be also used to provide LDA-like embeddings without being limited to using strictly less than $N_C$ dimensions. The S-LDA technique is compared to the LDA method using $N_C-1$ (the same dimensionality as the LDA method) and $2\times(N_C-1)$ dimensions. The results are shown in Table~\ref{table:sup-svm}. The S-LDA almost always outperforms the LDA method, especially when the training data are limited to 5 samples per class. Using more dimensions also provides some further accuracy gains. The statistical significance of the obtained results was also evaluated using a paired t-test. The null hypothesis (`There is no statistically significant difference between the LDA~($N_C-1$) and the S-LDA~($2 \times N_C-1$) methods.') is rejected ($p = 0.024 < a = 0.05$).

The LDA and the S-LDA methods are also compared using different number of dimensions in Figure \ref{fig:sup-svm-curves}. The \mbox{S-LDA} performs significantly better than the corresponding LDA method in most cases (except for a few experiments using the Yale and the KTH datasets). The accuracy of the \mbox{S-LDA} method usually converges when around $1.2-1.4 \times (N_C-1)$ dimensions are used.  Note that for the Yale dataset the weight of the orthonormality regularizer was set to $\alpha_P=0.0001$ instead of $\alpha_P=1$ (the latter led to suboptimal embeddings).

\subsection{SVM-based Analysis}

The proposed SVM-based projection (Subsection \ref{section:svm-based}) is evaluated in this Subsection. A linear SVM combined with the one-versus-one strategy is utilized. Again, the hyper-parameter of the SVM is chosen using cross-validation. The results are shown in Table \ref{table:svm}. In the last column the accuracy of the SVM classifier, which acts as the target for the \mbox{S-SVM-A} method, is shown. The \mbox{Raw + NCC } method refers to using the original representation of the data and an NCC classifier. The  \mbox{S-SVM-A} method is evaluated using $N_C$ and $2 \times N_C$ dimensions. The mean classification time per test sample is also reported next to each accuracy value.

The proposed technique (S-SVM-A + NCC) always outperforms the Raw + NCC method regardless the number of used dimensions. The SVM performs better than the proposed technique, which is expected since the S-SVM-A tries to perform SVM-like classification using only a linear projection. However, in one case (for KTH dataset) the proposed technique works slightly better than the SVM. The impact of the proposed technique on the accuracy (compared to an SVM) can be considered quite limited, especially if the simple nature of the used classifier and the performance benefits are considered (the classification time is reduced up to 3 orders of magnitude). The great speed benefits makes the proposed technique a great candidate for applications that involve real-time classification \cite{realtime}. 
\subsection{Out-of-Sample Extensions}

Finally, the ability of the SEF to provide out-of-sample extensions for the ISOMAP method \cite{isomap}, is evaluated using both linear (cS-ISOMAP) and kernel (cKS-ISOMAP) methods. The proposed methods are also compared to using Linear Regression (abbreviated as LR) and Kernel Ridge Regression (abbreviated as KR) respectively \cite{murphy2012machine}. For the ISOMAP technique the 30-nearest neighbor similarity graph (15-nn for the Yale dataset) is used and the target dimensionality is set to 10. In contrast to regression techniques, SEF is not restricted to learning out-of-sample extensions of the same dimensionality.  Therefore, the cS-ISOMAP and the cKS-ISOMAP techniques are evaluated using both 10 and 20 dimensions. Table \ref{table:oos-linear-svm} compares the linear out-of-sample extensions for the ISOMAP technique. The cS-ISOMAP provides better generalization ability than the Linear Regression. Also, learning higher-dimensional embeddings seems to increase classification accuracy. 

\begin{table}
\caption{Linear Out-Of-Sample Extensions (SVM Classification Accuracy)}	
\label{table:oos-linear-svm}
\centering
\begin{tabular}{l|ccc}
\textbf{dataset} &\textbf{LR(10)} &\textbf{cS-ISOMAP(10)} &\textbf{cS-ISOMAP(20)} \\ 
\hline 
\textbf{15-scene} &  $60.04 \pm 0.88$ &  $67.36 \pm 1.33$ &  $\mathbf{69.43 \pm 1.01}$ \\ 
\textbf{Corel} &  $35.71 \pm 0.58$ &  $38.64 \pm 0.74$ &  $\mathbf{42.21 \pm 0.55}$ \\ 
\textbf{MNIST} &  $85.33 \pm 0.46$ &  $85.87 \pm 0.39$ &  $\mathbf{88.83 \pm 0.36}$ \\ 
\textbf{Yale} &  $34.89 \pm 2.64$ &  $56.74 \pm 2.70$ &  $\mathbf{80.05 \pm 1.72}$ \\ 
\textbf{KTH} &  $63.82 \pm 1.61$ &  $84.19 \pm 0.14$ &  $\mathbf{84.82 \pm 1.23}$ \\ 
\textbf{20NG} &  $35.58 \pm 1.12$ &  $42.47 \pm 1.54$ &  $\mathbf{45.90 \pm 1.10}$ \\ 
\end{tabular}
\end{table}

Table \ref{table:oos-kernel-svm} also compares the kernel-based out-of-sample extensions. All the evaluated kernel techniques use an RBF kernel where the $\gamma$ parameter is chosen in such way to spread the similarities in the kernel matrix. Again, the cKS-ISOMAP performs better than the Kernel Ridge Regression, while using more dimensions seems to improve the generalization ability of the cKS-ISOMAP method.

\begin{table}
\caption{Kernel Out-Of-Sample Extensions (SVM Classification Accuracy)}	
\label{table:oos-kernel-svm}
\centering
	
\begin{tabular}{l|ccc}
\textbf{dataset} &\textbf{KR(10)} &\textbf{cKS-ISOMAP(10)} &\textbf{cKS-ISOMAP(20)} \\ 
\hline 
\textbf{15-scene} &  $62.50 \pm 0.56$ &  $64.18 \pm 0.74$ &  $\mathbf{68.07 \pm 0.76}$ \\ 
\textbf{Corel} &  $35.83 \pm 0.28$ &  $36.80 \pm 0.36$ &  $\mathbf{41.77 \pm 0.35}$ \\ 
\textbf{MNIST} &  $88.58 \pm 0.59$ &  $87.80 \pm 0.55$ &  $\mathbf{91.02 \pm 0.29}$ \\ 
\textbf{Yale} &  $44.02 \pm 3.59$ &  $35.49 \pm 2.36$ &  $\mathbf{59.32 \pm 1.61}$ \\ 
\textbf{KTH} &  $69.80 \pm 1.12$ &  $76.15 \pm 0.74$ &  $\mathbf{83.33 \pm 0.27}$ \\ 
\textbf{20NG} &  $44.63 \pm 0.83$ &  $41.09 \pm 0.82$ &  $\mathbf{48.25 \pm 0.92}$ \\ 
\end{tabular}
\end{table}

\section{Conclusion and Future Work}
\label{section:conclusions}
In this paper, the problem of Dimensionality Reduction was revisited from a new perspective. Instead of using second-order statistics to define the DR objective, as the vast majority of existing DR methods do, the target distribution was directly modeled using the notion of similarity. Based on this idea, a novel DR Framework, called Similarity Embedding Framework (SEF), was proposed. The proposed framework allows for expressing optimization targets inspired from existing techniques and overcoming some of their shortcomings, as well as, for easily deriving new novel techniques. The SEF was evaluated for a variety of classical tasks, such as performing supervised dimensionality reduction and providing out-of-of-sample extensions, as well as, new novel tasks, such as providing fast linear projections for complex techniques. For the evaluation six datasets from a diverse range of domains were used and it was demonstrated that the proposed methods can outperform many existing DR techniques.

It should be stressed again that the SEF allows for easily deriving new DR techniques. It is planned to publicly release a software package that implements the Similarity Embedding Framework to assist other researchers develop new techniques based on the proposed DR formulation. Some ideas for further research are briefly presented. The target similarity matrix can be set according to a clustering solution, to perform clustering-based DR. Preliminary experiments shows interesting results. Also, different classifiers can be considered to perform analysis similar to the (K)S-SVM-A technique. More complex targets, such as the distribution induced by the output of a deep convolutional network, e.g., VGG \cite{simonyan2014very}, can be set. This can provide a new way to perform neural network distillation \cite{hinton2015distilling}, using a simpler model. To further improve the performance of such model, more complex mapping functions, such as deep neural networks or convolutional networks, can be employed.

\ifCLASSOPTIONcaptionsoff
  \newpage
\fi



%

\bibliographystyle{IEEEtran}
\bibliography{bib}



%


\begin{IEEEbiography}[{\includegraphics[width=1in,height=1.25in,clip,keepaspectratio]{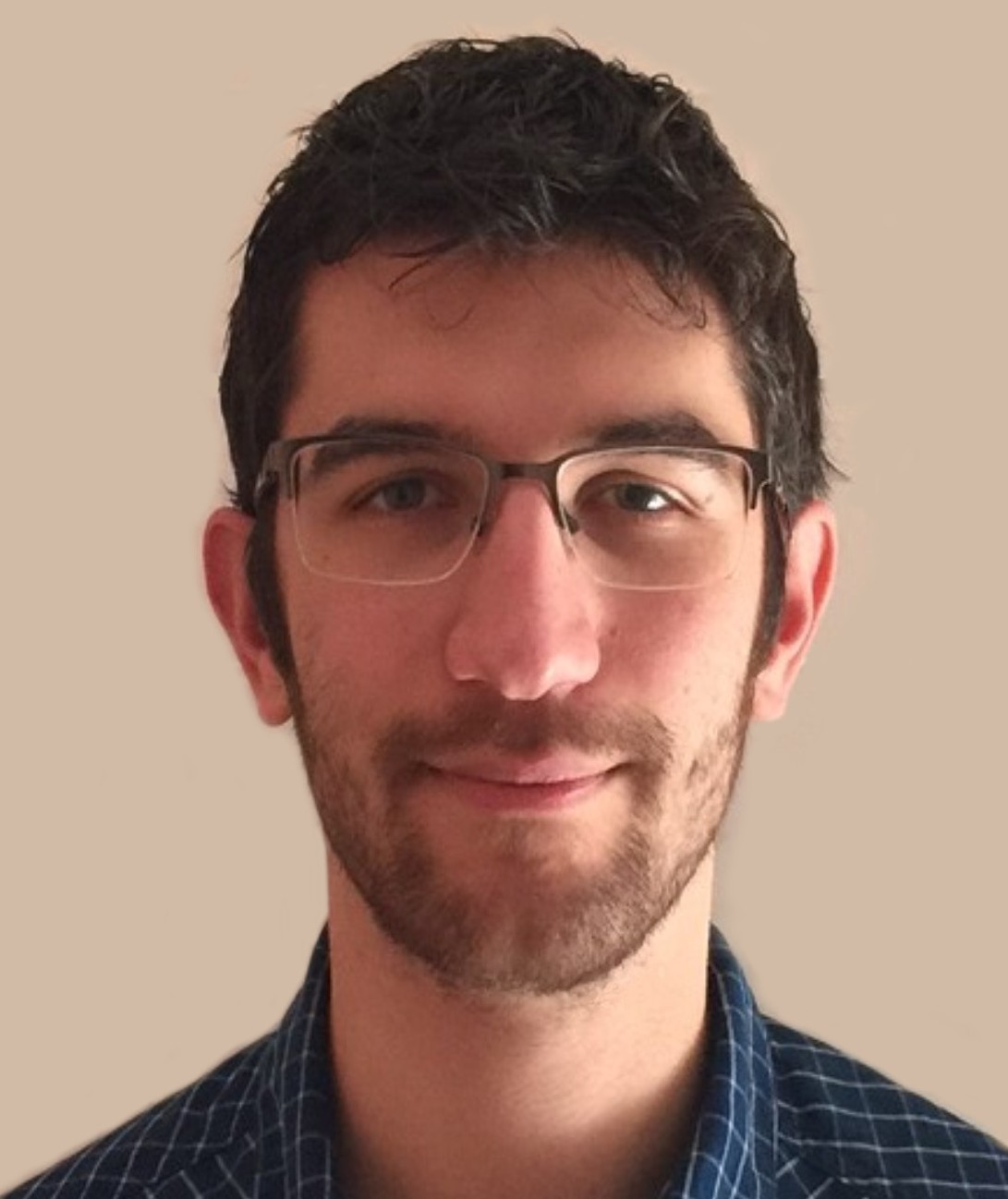}}]{Nikolaos Passalis}
	obtained his B.Sc. in informatics in 2013 and his M.Sc. in information systems in 2015 from Aristotle University of Thessaloniki, Greece. He is currently pursuing his Ph.D. studies in the Artificial Intelligence \& Information Analysis Laboratory in the Department of Informatics at the University of Thessaloniki.	His research interests include machine learning, computational intelligence and information retrieval. 
\end{IEEEbiography}

\begin{IEEEbiography}[{\includegraphics[width=1in,height=1.25in,clip,keepaspectratio]{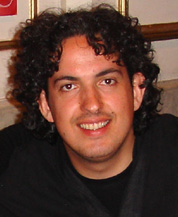}}]{Anastasios Tefas} received the B.Sc. in informatics in 1997 and the Ph.D. degree in informatics in 2002, both from the Aristotle University of Thessaloniki, Greece. Since 2017 he has been an Associate Professor at the Department of Informatics, Aristotle University of Thessaloniki. From 2008 to 2017, he was a Lecturer, Assistant Professor at the same University. From 2006 to 2008, he was an Assistant Professor at the Department of Information Management, Technological Institute of Kavala. From 2003 to 2004, he was a temporary lecturer in the Department of Informatics, University of Thessaloniki. From 1997 to 2002, he was a researcher and teaching assistant in the Department of Informatics, University of Thessaloniki. Dr. Tefas participated in 12 research projects financed by national and European funds. He has co-authored 72 journal papers, 177 papers in international conferences and contributed 8 chapters to edited books in his area of expertise. Over 3730 citations have been recorded to his publications and his H-index is 32 according to Google scholar. His current research interests include computational intelligence, deep learning, pattern recognition, statistical machine learning, digital signal and image analysis and retrieval and computer vision. 
\end{IEEEbiography}




\end{document}